\documentclass[a4,10pt,journal,compsoc]{IEEEtran}
\usepackage{listings}

\usepackage[colorinlistoftodos]{todonotes}
\usepackage[nocompress]{cite}
\usepackage{url}
\PassOptionsToPackage{hyphens}{url}
\usepackage{graphicx}
\usepackage{times}
\usepackage{natbib}
\usepackage{algorithm}
\usepackage{algpseudocode}
\algloopdefx{NFor}[1]{\textbf{for all} #1 \textbf{do}}
\usepackage{tikz}
\usetikzlibrary{positioning}% To get more advances positioning options
\usetikzlibrary{arrows}% To get more arrow heads
\usetikzlibrary{chains}

\title{{\bf Model-Based Deep Reinforcement Learning for
    High-Dimensional Problems,} {\em a Survey}} 
\author{Aske Plaat, Walter Kosters, Mike Preuss\\
  \small
  Leiden Institute of Advanced Computer Science}
\date{\today}

\begin{document}
\maketitle

\begin{abstract}

Deep reinforcement learning has shown remarkable success in the past
few years. Highly complex sequential 
decision making problems have  been solved in tasks
such as game
playing and robotics. Unfortunately, the sample complexity of most deep
reinforcement learning
methods is   high, precluding their use in some important
applications. Model-based reinforcement learning creates an explicit model  
of the environment dynamics  to reduce
the need for environment samples.
%as has been shown  in the literature on
%classic, non-deep, reinforcement learning.

Current deep learning methods use high-capacity networks to
solve high-dimensional problems. Unfortunately, high-capacity models typically require many
samples, negating the potential benefit of
lower sample complexity in model-based methods. A challenge for deep model-based methods is therefore to achieve
high predictive power while maintaining low sample complexity. 

In recent years, many model-based methods have been
introduced to address this challenge. In this paper, we survey the contemporary model-based
landscape. First we discuss definitions and relations to other
fields. % Then, we discuss applications in game playing and robotics.
We propose a
taxonomy based on three  approaches: using explicit planning on given transitions,
using explicit planning on learned transitions, and end-to-end
learning of both planning and transitions.
% five different approaches: model learning by backpropagation,
% hybrid imagination, planning networks, abstract models, and curriculum learning. 
%
We use these approaches to organize a comprehensive
overview of important recent developments such as latent models. We
describe  methods and benchmarks,  and we suggest
directions for future work for each of the approaches. Among promising
research directions are curriculum learning, uncertainty modeling, and
use of latent models for
transfer learning.
\end{abstract}

\section{Introduction}

%\todo{ add code pointers }
% \todo{Look at style of intro. This reads like a text book explaining
%   things that are insultingly obvious to the intended audience. Look
%   at Deisenroth/Kober/Stone surveys to look at their style of intro}
% \todo{They describe the problem, to which the topic is a solution}
% \todo{Stonne:}
% \todo{problem: sample eff of RL. solution: model based RL. problem:
%   hi-dim good models. solution: we will see vrious approaches}
% \todo{MBRL is important because: great deep advances, can solve
%   further challenges. Recent workshops?}
% \todo{Kober: is in tutorial style like mine. Deisenroth starts
%   tutorial style, but goes deep quickly. Kaelbling too. They all go
%   deep quicker than I do}
% \todo{Hmarick: want to show analogy with cognition}

Deep reinforcement learning has shown remarkable successes in the past
few years. Applications in game playing and robotics  have shown the
power of this
paradigm with applications such as learning to play Go from scratch or flying an acrobatic model helicopter~\citep{mnih2015human,silver2016mastering,abbeel2007application}. 
Reinforcement learning uses an environment from which training
data is sampled; in contrast to supervised learning it does not need 
a large database of pre-labeled training data. This opens up many
applications for machine learning for which no such database
exists. Unfortunately, however, for most interesting applications 
many samples from the environment are necessary, and the computational cost of learning
is prohibitive, a problem that is common in deep
learning~\citep{lecun2015deep}. Achieving faster learning is a
major goal of much current research. Many promising approaches are
tried, among them
metalearning~\citep{hospedales2020meta,huisman2020deep}, transfer
learning~\citep{pan2010survey}, curriculum
learning~\citep{narvekar2020curriculum} and zero-shot
learning~\citep{xian2017zero}. The current paper focuses on
model-based methods in deep reinforcement learning.

Model-based methods can reduce  sample complexity. In contrast to
model-free methods that sample at will from the environment, model-based methods build up a dynamics model of the
environment as they sample. By using this dynamics model for policy updates, the number of
necessary samples can be reduced
substantially~\citep{sutton1991dyna}. Especially in robotics sample-efficiency is important (in games environment
samples can often be generated more cheaply).

The success of the model-based
approach hinges critically on the quality of the predictions of the
dynamics model, and here the prevalence of deep learning presents a
challenge~\citep{talvitie2015agnostic}. Modeling the dynamics of high dimensional 
problems usually requires high capacity networks that, unfortunately, require many samples
for training to achieve high generalization while preventing overfitting, potentially undoing the sample efficiency gains of
model-based methods. Thus, the problem statement of the methods in this survey
is {\em how  to train a high-capacity dynamics model with
high predictive power and low sample complexity}.

In addition to promising better sample efficiency than model-free
methods, there is another reason for the interest in  model-based
methods for deep learning. Many problems in reinforcement learning are
sequential decision problems, and learning the transition function is
a natural way 
% , and the ``flat'' approach of learning
% state/action pairs may not be the most efficient approach for
% learning a the policy. The policy consists of a sequence of decision
% points. 
of capturing the core of long and complex decision sequences. This is
what is
called a forward model in game AI~\citep{risi2020chess,torrado2018deep}.
% The
% transition model approach
% is most likely  more efficient than the model-free approach of
% blind sampling of state/action pairs.
% It can be argued that for
% sequential decision problems, learning the transition function is a
% more important goal than the policy,
When a  good transition
function of the domain is present, then new,  unseen, problems can be
solved efficiently.  Hence,
model-based reinforcement learning may contribute to efficient transfer
learning. % and provides better generalization for complex sequential decision problems.

% Recent
% advances in deep learning %and in model-based reinforcement learning
% have created a flurry of work in model-based {\em deep} reinforcement
% learning
The  contribution of this survey is to give an in-depth
overview of recent methods
 for model-based deep reinforcement learning. We describe
methods that use
(1)  explicit planning on given transitions,
(2)  explicit planning on a learned transition model, and
(3) end-to-end learning of both planning and transitions.
%model learning by backpropagation,
%hybrid imagination,
%planning networks, abstract models, and curriculum learning.
For each approach
future directions are listed (specifically: latent models, uncertainty
modeling, curriculum learning and
multi-agent benchmarks). 
% Team collaboration, negotiation, and competition can be tested in
% multi-agent real time stategy games, and in robotics end-to-end
% learning of camera-based
% robot manipulation tasks provides important challenges. 
% Some
                                % of the reviews include some work on
                                % deep model-based 
% reinforcement learning. We believe that an overview of recent
% developments is useful.

Many research papers have
been published recently, and the field of model-based deep
reinforcement learning is advancing rapidly. %%  Different approaches are
%% described in the papers, for different applications.
The papers in
this survey are selected on recency and impact on the field, for different
applications, highlighting relationships between papers. Since our
focus is  on recent work, some of the references are to  preprints in arXiv
(of reputable groups). Excellent works with necessary background
information exist for  reinforcement
learning~\citep{sutton2018introduction}, deep learning~\citep{goodfellow2016deep}, machine
learning~\citep{bishop2006pattern}, and artificial
intelligence~\citep{russell2016artificial}.
As we mentioned, the
main purpose of the current survey is to focus on deep learning
methods, with high-capacity models.
Previous surveys provide
an overview of the uses of classic (non-deep) model-based
methods~\citep{deisenroth2013survey,kober2013reinforcement,kaelbling1996reinforcement}.
Other relevant surveys into model-based reinforcement learning
are~\citep{justesen2019deep,polydoros2017survey,hui2018model,wang2019benchmarking,ccalicsir2019model,moerland2020model}. 

The remainder of this survey is structured as
follows. Section~\ref{sec:rl} provides necessary background and a familiar
formalism of reinforcement learning.  Section~\ref{sec:mbrl} then
surveys  recent papers in the field of  model-based deep
reinforcement learning. Section~\ref{sec:bench}
introduces the main benchmarks of the field. Section~\ref{sec:dis} provides a discussion
reflecting on the different approaches and provides   open
problems and future work. Section~\ref{sec:con} concludes the survey.

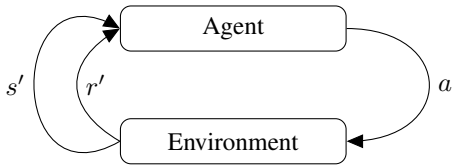
\begin{figure}[h]
  \begin{center}
    \begin{tikzpicture}[>=triangle 45,
  desc/.style={
		scale=1.0,
		rectangle,
		rounded corners,
		draw=black, 
		}]

  \node [desc,minimum width=3cm,minimum height=0.6cm] (tm) at   (0,0.5) {Environment};
  \node [desc,minimum width=3cm,minimum height=0.6cm] (pol) at   (0,2) {Agent};
  \draw (tm.west) edge[->,in=210,out=150,looseness=1.5] node[right] {$r'$} (pol.west);
  \draw (tm.west) edge[->,in=150,out=210,looseness=3] node[left] {$s'$} (pol.west);
  \draw (pol.east) edge[->,out=0,in=0,looseness=2.5] node[right] {$a$} (tm.east);

\end{tikzpicture}
  \end{center}
  \caption{Reinforcement Learning: Agent Acting on Environment, that
    provides new State and Reward to the Agent}\label{fig:agent}
\end{figure}

\section{Background}\label{sec:rl}
% RL needs model or env
Reinforcement learning does not assume the presence of a
database, as supervised learning does. Instead, it derives the ground
truth from  an internal model or from an external environment that can 
be queried by the learning agent, see Figure~\ref{fig:agent}. The environment
provides a new state $s'$ and its reward $r'$ (label) for every
action $a$ that the agent tries in a certain state $s$~\citep{sutton2018introduction}. In this way, as
many action-reward pairs can be 
generated as needed, without  a large hand-labeled database. Also, we
can learn behavior beyond that what a supervisor prepared for us to learn.

As  so much of artificial intelligence, reinforcement learning draws
inspiration from principles of human and 
animal learning~\citep{hamrick2019analogues,kahneman2011thinking}.  In psychology, learning is studied
as behaviorial  adaptation, as a result of reinforcing
reward and punishment. Publications in
artificial intelligence sometimes explicitly reference analogies in
how learning in the two fields  is described~\citep{anthony2017thinking,duan2016rl,weng2020meta}. 

% Unsupervised learning learns
% a function without labeled examples or an
% environment, by looking at inherent features of data, such as
% distance. We will encounter unsupervised learning in 
% our discussion on abstract models in variational autoencoders and generative adversarial
% networks in Section~\ref{sec:abstract}.

Supervised learning frequently studies 
regression and classification problems. In reinforcement learning most
problems are decision and control problems.
%An action must be taken,
%but which?
Often problems are sequential decision problems,
in which a goal is reached after a sequence of decisions are
taken (behavior). In sequential decision making, the dynamics of the world are
taken into consideration. Sequential decision making is a
step-by-step approach  in which earlier decisions influence later
decisions.
Before we continue, let us formalize key concepts in reinforcement learning.

\subsection{Formalizing Reinforcement Learning}\label{sec:mdp}
% mdp, policy, value

Reinforcement learning problems are often modeled formally as a Markov
Decision Process (MDP). First we  introduce the basics: state,
action, transition and reward. Then we introduce policy and
value. Finally, we define model-based and model-free solution approaches.

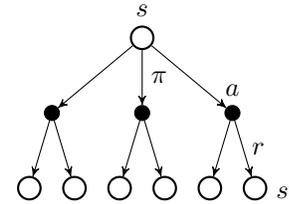
\begin{figure}[t]
\begin{center}
  \tikzset{
  treenode/.style = {align=center, inner sep=0pt, text centered,
    font=\sffamily},
  arn_n/.style = {treenode, circle, white, draw=black,
    fill=black, text width=2mm},% arbre rouge noir, noeud noir
  arn_r/.style = {treenode, circle, black, draw=black, 
    text width=3mm, thick}
}

\begin{tikzpicture}[->,>=stealth',level/.style={sibling distance = 1.2cm/#1,
  level distance = 1cm}] 
\node [arn_r,label=above:{$s$}] {}
    child{ node [arn_n] {} 
            child{ node [arn_r] {} 
            }
            child{ node [arn_r] {}
            }                            
    }
    child{ node [arn_n] {} 
            child{ node [arn_r] {}
            }            
            child{ node [arn_r] {}
            }
            edge from parent node[right] {$\pi$} 
    }
    child{ node [arn_n,label=above:{$a$}] {}
            child{ node [arn_r] {} 
            }
            child{ node [arn_r,label=right:{$s'$}] {}  edge from parent node[right] {$r$} 
            }
    }
;  
\end{tikzpicture}
\caption{Backup Diagram~\citep{sutton2018introduction}. 
Maximizing  the reward for state $s$ is done by following the {\em transition}
function to find the next state $s'$. Note that the policy $\pi(s,a)$ tells
the first half of this story, going from $s \rightarrow a$; the
transition function $T_a(s,s^\prime)$ completes the  story, going from $s \rightarrow
s^\prime$ (via $a$).}\label{fig:rltree}
\end{center}
\end{figure}

A Markov Decision Process is a 4-tuple $(S, A, T_a, R_a)$ where
$S$ is a finite set of states,
$A$ is a finite set of actions; $A_s \subseteq A$ is the set of actions available
from state $s$. Furthermore, $T_a$ is the transition function: $T_a(s,s')$ is
the probability that action $a$ 
in state $s$ at time $t$ will lead to state $s^\prime$ at time
$t+1$. Finally, $R_a(s,s^\prime)$ is the immediate reward received after transitioning
from state $s$ to state $s^\prime$ due to action $a$. The goal of an
MDP is to find the best decision, or action, in all states $s \in
S$.

The goal of reinforcement learning is  to find the optimal policy $a=\pi^\star(s)$,  which is the
function that gives
the best action $a$ in all states  $s\in S$. The policy contains the
actions of the answer to a sequential
decision problem: a step-by-step prescription of which action must be
taken in which state, in order to maximize reward for any given
state. This policy can be found
directly---model-free---or with the help of a transition model---model-based.
Figure~\ref{fig:rltree} shows a diagram of the transitions.
More formally, the goal of an MDP is to find  policy $\pi(s)$ that
chooses an action in state 
$s$ that will maximize the reward. This value $V$ is the expected sum of 
 future rewards $V^\pi(s)=E(\sum_{t=0}^\infty \gamma^t
R_{\pi(s_t)}(s_t, s_{t+1}))$ that are discounted with parameter
$\gamma$ over $t$ time periods, with $s = s_0$.  The function $V^\pi(s)$ is called the value function of the
state. In deep learning  the policy $\pi$ is determined by the parameters $\theta$
(or weights) of a neural
network, and the parameterized policy  is  written as $\pi_\theta$.

There are algorithms to compute the policy $\pi$ directly, and there are
algorithms that first compute this
function $V^\pi(s)$. For stochastic problems often  direct policy methods work
best, for deterministic problems the value-methods are most often
used~\citep{kaelbling1996reinforcement}. (A third, quite popular, approach combines the
best of value and policy methods: actor-critic~\citep{sutton2018introduction,konda2000actor,mnih2016asynchronous}.) In classical, table-based,
reinforcement learning there is a close relation 
between policy and value, since the 
best action of a state leads to both the best policy and the best
value, and finding the other can  usually be done  with a simple
lookup. When the value and policy function are approximated, for
example  with a
neural network, then this relation becomes weaker, and many advanced
policy and value algorithms have been devised for deep reinforcement learning.

Value function algorithms  calculate the state-action value
$Q^\pi(s,a)$. This $Q$-function
gives the expected sum of discounted rewards when following action $a$
in state $s$, and then afterwards policy $\pi$. The value $V(s)$  is
the maximum of the $Q(s,a)$-values of that state. The optimal 
value-function is denoted as $V^\star(s)$. The optimal policy
can be found by recursively
choosing the argmax action with $Q(s,a)=V^\star(s)$  in 
each state. 

To find the policy by planning, models for $T$ and
$R$ must be known. When they are not known, an environment is assumed
to be present for the 
agent to query in order to get the necessary
reinforcing information, see Figure~\ref{fig:agent},
after~\cite{sutton2018introduction}. The samples can be used to build
the model of $T$ and $R$ (model-based reinforcement learning) or they
can be used to find the policy without first building the model (direct or
model-free reinforcement learning). When sampling, the
environment is in a known state $s$, and 
the agent  chooses an action $a$ which it transmits to the
environment, that responds with a new state $s^\prime$ and the
corresponding reward value $r'=R_a(s, s^\prime)$.

% In reinforcement learning, 
% either the transition and reward model  is explicitly known by the
% agent---model-based---or the models  are implicit in the environment---model-free. $T$ and $R$ or
% the environment samples are used
% by solution algorithms to find the optimal policy $\pi^\star$ and the
% optimal state value $V^\star$.

The literature provides many solution
algorithms. We now very briefly discuss classical planning and model-free approaches,
before we continue to survey  model-based algorithms in more depth in
the next section.

\subsection{Planning}\label{sec:planning}
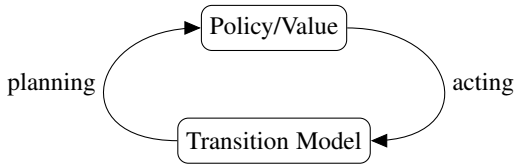
\begin{figure}[t]
  \begin{center}
    \begin{tikzpicture}[>=triangle 45,
  desc/.style={
		scale=1.0,
		rectangle,
		rounded corners,
		draw=black, 
		}]

  \node [desc,minimum height=0.6cm] (tm) at   (0,0.5) {Transition Model};
  \node [desc,minimum height=0.6cm] (pol) at   (0,2) {Policy/Value};
  \draw (tm.west) edge[->,in=180,out=180,looseness=2.5] node[left] {planning} (pol.west);
  \draw (pol.east) edge[->,out=0,in=0,looseness=2.5] node[right] {acting} (tm.east);

\end{tikzpicture}
    \caption{Planning}\label{fig:plan}
  \end{center}
\end{figure}

% planning: model is available in agent
% no sampling
Planning algorithms use the transition model to find the optimal policy, by
selecting actions in states, looking ahead, and backing up reward
values, see Figure~\ref{fig:rltree} and Figure~\ref{fig:plan}.

\begin{algorithm}
  \caption{Value Iteration}\label{lst:vi}
    \begin{algorithmic}
      \State Initialize $V(s)$ to arbitrary values
      \Repeat 
      \ForAll{$s$}
      \ForAll{$a$}
      \State $Q[s,a] = \sum_{s'} T_a(s,s')(R_a(s,s') + \gamma V(s'))$
      \EndFor
      \State $V[s] = \max_a(Q[s,a])$
      \EndFor  
      \Until V converges
      \State return V
    \end{algorithmic}
\end{algorithm}

In planning algorithms,  the agent has access to an explicit transition
and reward model. In the deterministic case the transition model provides the next state for each of the
possible actions  in  the states, it is a function $s^\prime =
T_a(s)$. In the  stochastic case, it provides the probability
 distribution $T_a(s, s^\prime)$. The reward
model provides the immediate reward  for transitioning from state $s$
to state $s^\prime$ after taking action $a$. Figure~\ref{fig:rltree}
provides a backup diagram for the  transition and 
reward function. The transition function moves downward in the diagram from
state $s$ to $s'$, and
the reward value goes upward in the diagram, backing up the value from
the child state to
the parent state. The  transition function follows policy $\pi$ with
action $a$, after which state $s^\prime$ is chosen with probability
$p$, yielding reward $r'$.  The policy function $\pi(s,a)$ concerns the top layer of the diagram, from $s$ to $a$. The transition function $T_a(s,s^\prime)$ covers both layers, from $s$ to $s^\prime$. In some domains, such as chess, there is a
single deterministic state $s^\prime$ for each action $a$. Here each move
leads to a single board position, simplifying the backup diagram.

Together, the transition and reward functions implicitly define a space of
 states that can be searched for the optimal policy $\pi^\star$ and
 value $V^\star$.

 The most basic form of planning is Bellman's dynamic programming~\citep{bellman1957dynamic}, a recursive
 traversal of the state and action space. Value iteration is a well-known, 
 very basic, dynamic programming
 method. The pseudo-code for value iteration is shown in
 Algorithm~\ref{lst:vi}~\citep{alpaydin2020introduction}. It traverses
 all actions in all states, 
 computing the value of the entire state space. 
 
Many planning algorithms have been devised to efficiently generate and
traverse
state spaces, such as % basic depth-first search, breadth-first
% search, and well-known AI algorithms such as
(depth-limited) A*,  alpha-beta and
Monte Carlo Tree Search (MCTS)~\citep{hart1968formal,pearl1984heuristics,korf1985depth,plaat1996best,browne2012survey,moerland2018a0c,moerland2020second}.

Planning algorithms originated from exact, table-based,
algorithms~\citep{sutton2018introduction} that fit in the symbolic AI
tradition. For  planning it is  relevant to know  how much of the
state space must be traversed to find the optimal policy. When state spaces are too large to search fully, 
 deep function approximation algorithms can be used  to approximate the
optimal policy and
value~\citep{sutton2018introduction,plaat2020learning}.

Planning is sample-efficient in the sense that, when the agent has a model, a policy can be found 
without %ever sampling  from the environment. All planning can take place without
interaction with the environment. Sampling may be
costly, and sample efficiency is an important concept 
in reinforcement learning.

A sampling action taken in an environment is irreversible, since state changes
of the environment  can not be undone by the agent. In
contrast, a planning
action taken in a transition model is
reversible~\citep{moerland2020framework}. A planning agent can 
backtrack, a sampling agent cannot. Sampling  finds
local optima easily. For finding global optima the ability to backtrack
out of a local optimum is useful, which is an advantage for
model-based planning methods.

Note, however, that there are two ways of finding dynamics models. In some problems, the
transition and reward models are given by the problem,
such as in games, where the move rules are known, as in Go
and chess.  Here the
dynamics models follow the 
problem perfectly, and many steps can be planned accurately into the
future without problem, out-performing model-free sampling. In
other problems the dynamics model must be 
learned from sampling the environment. Here the model will not be
perfect, and will contain errors and biases.
Planning far ahead will only work when  the agent has a $T$ and $R$ model of
sufficient quality. % to 
% determine the best policy. 
% Of puzzles and perfect information games such as Go and chess 
% the transition and reward rules for
% moving and win/loss are known. When they are not known, then it may be
% possible to sample the environment to learn the models from data (see
% Section~\ref{sec:mbrl}), 
% , no sampling with the environment is
% necessary to learn the models. Efficient planning algorithms exist to
% traverse the state space. Even large state spaces have been searched
% fully~\cite{schaeffer2007checkers}.
% or a limited number of steps
% can be planned ahead.
With learned models, it may be more difficult for
model-based planning to achieve the performance of model-free sampling.

\begin{figure}[t]
  \begin{center}
    \begin{tikzpicture}[>=triangle 45,
  desc/.style={
		scale=1.0,
		rectangle,
		rounded corners,
		draw=black, 
		}]

  \node [desc,minimum height=0.6cm] (tm) at   (0,0.5) {Environment};
  \node [desc,minimum height=0.6cm] (pol) at   (0,2) {Policy/Value};
  \draw (tm.west) edge[->,in=180,out=180,looseness=2.5] node[left]
  {learning} (pol.west);
  \draw (pol.east) edge[->,out=0,in=0,looseness=2.5] node[right] {acting} (tm.east);

\end{tikzpicture}
    \caption{Model-Free Learning}\label{fig:free}
  \end{center}
\end{figure}
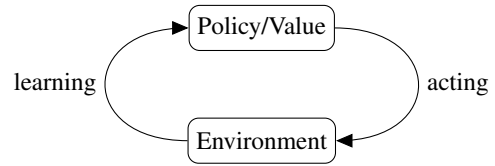

\subsection{Model-Free }
% learning: model is available in env
% not sample efficient, since no model of transitions is present, and 
When the transition
or reward model  are not available to 
the agent, then the policy and value function  have to be learned through querying the
environment. Learning the policy or value
function without a model,  
through sampling the environment, is called model-free learning, see
Figure~\ref{fig:free}.

Recall that the policy is  a mapping of states to best actions. Each time when a new reward is returned by the environment
the policy can be improved: the best action for the state is
updated to reflect the new information.
Algorithm~\ref{lst:free} shows the simple high-level steps of model-free
reinforcement learning (later on the algorithms become more elaborate).

\begin{algorithm}
    \caption{Model-Free Learning}\label{lst:free}
    \begin{algorithmic}
      \Repeat
      \State Sample env $E$ to generate data $D=(s, a, r', s')$  
      \State Use $D$ to update policy $\pi(s, a)$
      \Until $\pi$ converges
    \end{algorithmic}
\end{algorithm}

Model-free reinforcement learning is the most basic form of
reinforcement learning. It has been successfully applied to a range of
challenging
problems~\citep{deisenroth2013survey,kober2013reinforcement}. 
In model-free reinforcement learning a policy is learned from the
ground up through interactions (samples)
with the environment.

The main goal of model-free learning is to achieve good generalization: to
achieve high accuracy on test problems not seen during training. A
secondary goal is to do so with good sample efficiency: to need as
few environment samples as possible for good generalization. 

Model-free learning is essentially blind, and
learning the policy and value takes many  samples.  A
well-known model-free reinforcement learning algorithm is 
Q-learning~\citep{watkins1989learning}.  Algorithms such as
Q-learning can be used in a classical
table-based setting. Deep
neural networks have also been used with  success in model-free
learning, in domains in which samples can be
generated cheaply and quickly, such as in Atari video
games~\citep{mnih2015human}. Deep model-free algorithms such as Deep
Q-Network (DQN)~\citep{mnih2013playing} and
Proximal Policy Optimzation (PPO)~\citep{schulman2017proximal} have
become quite popular. PPO is an algorithm that computes the policy
directly, DQN finds the value function first (Section~\ref{sec:mdp}).

Although risky, an advantage of flying blind is the absence of
bias. Model-free reinforcement learning 
can find global optima without being distracted by
a biased model (it has no model). Learned models in model-based
reinforcement learning may introduce bias, and model-based methods may not
always be able to find as good results as model-free can (although it
does find the biased results with fewer samples).

% Note that we did not mention bias in the previous section on planning,
% since the transition and reward models in the planning situation are
% assumed to be given by the domain, and are in that sense perfect.  In
% model-based learning  the imperfect model may introduce bias.

Let us look at the cost of our methods.
% Model-free reinforcement learning has been extentended to large neural
% network policy an value functions, making end-to-end training of
% complex tasks possible~\citep{mnih2015human,schulman2015trust}.
% In planning, when the state space is very large, traversing
% the entire space is costly.
Interaction with the environment %, on the other hand,
can  be costly. Especially when the
environment involves the real world, such as in real robot-interaction,
then sampling should  be minimized, for reasons of cost, and
to prevent wear of the robot arm.
In virtual environments on the other hand, model-free approaches have been quite
successful, as we have noted in Atari and other game play~\citep{mnih2015human}.

% To make the bridge to model-based reinforcement learning, we 
% note that in model-free reinforcement learning a policy or a value
% model are learned. Policy models associate actions with states, and
% value models associate values with states. The fact that transition models
% are like policy models---they 
% also associate actions with states---and reward models are like value
% models, has created an interest in model-based reinforcement learning,
% in which planning and learning work together in using the learned
% policy and value models for transitions and rewards, in order to
% improve sample-efficiency by using planning using the models that have
% been created by learning from the environment.
A good overview of model-free
reinforcement learning can be found in~\citep{ccalicsir2019model,sutton2018introduction,kaelbling1996reinforcement}.

\subsection{Model-Based }
% combine planning and learning: model is learned in agent out of
% interaction with env and then used in agent
It is now time to look at model-based reinforcement learning, a method
that learns the policy and
value  in a different way than by sampling the
environment directly. Recall that the environment
samples return $(s',r')$ pairs, when the agent selects action $a$ in
state $s$. Therefore all information is present to learn the
transition model $T_a(s,s')$ and the reward model $R_a(s,s')$, for
example by supervised learning. When no transition model is given by
the problem, then the model can be learned by sampling the
environment, and  be 
used 
with planning to update the policy and value as often as we like. This alternative approach
of finding the policy and the value is called model-based learning. 

If the model is given, then no environment samples are needed and 
model-based methods are  more sample efficient. But if the
model is not given, why would we want to go this convoluted model-and-planning route, if the samples
can teach us the optimal policy and value directly? The reason is that
the convoluted route may be more sample efficient. % We note that

When
the complexity of learning the transition/reward model is smaller than the
complexity of learning the policy model directly, and planning is fast,
then the model-based route   can  be more efficient.
In model-free learning
a sample is used once to optimize the policy, and then thrown away, in model-based learning
the sample is used to learn a transition model, which can then be used many
times in planning to optimize the policy. The sample is used more efficiently.

% Most modern model-free methods use deep learning to approximate their
% policy and value functions, and hence in this survey we study deep
% model-based reinforcement learning methods.

% Model-based reinforcement learning is a data-efficient approach to
% learning control tasks but is difficult to utilize in domains with
% hihg-dimensional or complex observation such as images.

The recent successes in deep
learning caused much interest and 
progress in deep model-free learning.
Many of the deep function approximation methods that have been so
successful in  supervised learning~\citep{goodfellow2016deep,lecun2015deep} can also be used in model-free
reinforcement learning for approximating the policy and value
function. 

However, there are reasons for interest in model-based methods as
well. Many real world problems are long and complex sequential decision making problems, and
we are now seeing efforts to make progress in model-based methods. 
Furthermore, the interest in lifelong learning stimulates interest in
model-based learning~\citep{silver2013lifelong}. 
Model-based reinforcement learning is close to human and animal
learning, in that all new knowledge is interpreted in the context of
existing knowledge. The dynamics model is used to process and interpret new samples, in
contrast to model-free learning, where all samples, old and new, are
treated alike, and are not interpreted using the knowledge that has
been accumulated so far in the model. 

% methods in {\bf We will group them into
%   three classes: imagination, derivatives, and search-networks. Welke
%   groepering zal ik nemen?? derivatives, point, distributions, search networks,
%   observations/abstract, selfplay. is niet echt geordend. een beetje
%   een rommeltje. dyna, self-play,networks, abstract? Dyna->Self-play
%   is the step from model in agent to model in environment. Network and
% Abstract are important structural/architectural innovations. Actually,
% these are the three architectural innnovations. The others are
% non-architectural imporvmenete (distributional, derivatives. Or just
% make a nice list of these improvements and their description,
% afterwards, in discussion?}

% An advantage of model-based over model-free learning is that the model
% is able to transfer knowledge over
% different tasks, and thus achieve transfer learning.

After these introductory words, we are now ready to take a deeper look
into recent concrete deep model-based reinforcement learning methods.

\section{Survey of Model-Based Deep Reinforcement Learning}\label{sec:mbrl}

The success of model-based reinforcement learning depends  on the
quality of the dynamics model.
% Only good  models will lead to better or
% faster decisions than model-free approaches.
The model is typically used by planning
algorithms for multiple sequential predictions, and errors in
predictions accumulate quickly. %, and then model-free approaches will be
%preferred. %% For traditional, table-based methods, 
%% solving low-dimensional problems, model accuracy can often be achieved
%% with few samples.  Modeling the
%% dynamics of high dimensional problems requires high capacity networks
%% that require many samples for training. This high sample compexity may
%% offset potential sample efficiency gains of the model-based
%% approach. Therefore, the main 
%% challenge for \emph{deep} model-based reinforcement learning is to
%% find a method to
%% train  high capacity dynamics models that achieve  high 
%% predictive power using few samples. 
%
%% Recent years have seen a large activity yielding a large  variety in the
%% methods that have been tried. 
We group the methods  in three main
approaches. First the  transitions  are given and used by explicit planning, second the transitions
are learned and used by explicit planning, and third both transitions
and planning are learned end-to-end:

\begin{enumerate}
\item {\bf Explicit Planning on Given Transitions}\\
First,  we discuss methods for problems that give us clear
transition rules. In this case transition models are perfect, and classical,
explicit, planning methods are used to optimize the value and policy
functions of large state spaces. Recently, large and complex problems have been solved in
two-agent games using
self-learning methods that give rise to curriculum learning. Curriculum learning has also been used in single
agent problems.

\item {\bf Explicit Planning on Learned Transitions}\\
Second, we discuss methods for problems where no clear rules exist, and the transition model must be learned
from sampling the environment. (The transitions are again used with
conventional planning methods.) The environment samples allow learning
by backpropagation 
of high-capacity models. It is important that
the model has as few errors as possible. Uncertainty modeling and
limited lookahead can reduce the impact of prediction errors.

\item {\bf End-to-end Learning of Planning and Transitions}\\
Third, we discuss the situation where both the transition model and
the planning algorithm are learned from the samples, end-to-end.
A  neural network can be used in a way that it performs the actual
steps of certain planners, in addition to learning the
transitions from the samples, as before. The model-based algorithm is
learned fully end-to-end. 
A drawback of
this approach is the tight connection between network architecture
and problem type, limiting its applicabily. This drawback can be resolved
with the use of latent models, see below.  %% Related to this 
%% approach is the use of planning algorithms in neurons in the
%% network.
\end{enumerate}
In addition to the three main approaches, we now  discuss two
orthogonal approaches. These can be used to improve
performance of the three main approaches. They are the hybrid
imagination idea from Sutton's Dyna~\citep{sutton1991dyna}, and abstract, or latent, models.
\begin{itemize}
\item {\bf Hybrid Model-Free/Model-Based Imagination}\\
We first mention a sub-approach where environment samples are not only used to train
the transition model, but also to train the policy function directly, just as in
model-free learning. This hybrid approach thus combines
model-based and model-free learning. It is also called 
\emph{imagination} because the looking ahead  with the
dynamics model resembles simulating or imagining environment samples outside the
real environment. In this approach the imagined, or planned,  ``samples'' augment
the real (environment) samples. This augmentation reduces sample
complexity of model-free methods.

\item {\bf Latent Models}\\
Next, we discuss a sub-approach where the learned dynamics model is
split into several lower-capacity, specialized, latent
models. % of the dynamics of the environment are used for
% learning the policy.
These  latent models are then used with
planning or imagination to find the policy.  Latent models have been
used with and without end-to-end model training and with and without
imagination. Latent models thus build on and improve the previous approaches.
\end{itemize}
\begin{table*}[h]
  \begin{center}\footnotesize
    \begin{tabular}{llllccl}
      {\em Approach}&{\em Name}&{\em Learning}&{\em Planning}&{\em
                                                               Hybrid}&{\em
                                                                      Latent}&{\em
                                                                                 Application}\\
& & & & {\em Imagination}&{\em Models}& \\
      \hline\hline
%      Model Learning&PILCO~\citep{deisenroth2011pilco}&Gaussian Process&Cartpole\\
 %             &iLQG~\citep{tassa2012synthesis}&Least-Squares&Humanoid\\
  %            &GPS~\citep{levine2014learning}&Trajectory Optim&Humanoid\\
   %           &SVG~\citep{heess2015learning}&Value Gradients&Cheetah\\
    %          & PETS \citep{chua2018deep}&Ensembles&Cheetah\\
     %         & Ensemble TRPO \citep{kurutach2018model}&Ensembles&Cheetah\\
      %              &Visual Foresight \citep{finn2017deep}&MPC&Grasping\\
      Explicit Planning&TD-Gammon \citep{tesauro1995td}&Fully connected net&Alpha-beta&-&-&Backgammon\\
      Given Transitions   &Expert Iteration \citep{anthony2017thinking}&Policy/Value CNN&MCTS&-&-&Hex\\
      (Sect.~\ref{sec:selfplay})        &Alpha(Go) Zero \citep{silver2017mastering}&Policy/Value ResNet&MCTS&-&-&Go/chess/shogi\\
              &Single Agent \citep{feng2020solving}&Resnet&MCTS&-&-&Sokoban\\
      \hline
      Explicit Planning&PILCO~\citep{deisenroth2011pilco}&Gaussian Processes& Gradient based &-&-&Pendulum\\
Learned Transitions&      iLQG~\citep{tassa2012synthesis}& Quadratic Non-linear &MPC&-&-& Humanoid\\
 (Sect.~\ref{sec:foresight}) &     GPS~\citep{levine2014learning}& iLQG& Trajectory&-&-&Swimmer\\
  &    SVG~\citep{heess2015learning}&Value Gradients& Trajectory &-&-& Swimmer\\
   &   PETS \citep{chua2018deep}& Uncertainty Ensemble&MPC&-&-& Cheetah\\
%    &  ME-TRPO\citep{kurutach2018model}&Ensemble&Sample&-&-&Cheetah\\
     & Visual Foresight \citep{finn2017deep}& Video Prediction&MPC&-&-&Manipulation\\
%              &Local Model~\citep{gu2016continuous}&Local Q-learning&Cheetah\\
%              &MVE \citep{feinberg2018model}       &Uncertainty&Cheetah\\
%              &Meta-Policy \citep{clavera2018model}&Ensembles&Cheetah\\
%              &GATS \citep{azizzadenesheli2018surprising}&Pix2Pix+MCTS&Cheetah\\
%      &Policy Optim \citep{janner2019trust}&Short rollouts&Cheetah\\
      & Local Model~\citep{gu2016continuous}& Quadratic Non-linear& Short rollouts &+&-& Cheetah\\
      &MVE  \citep{feinberg2018model}  & Samples& Short rollouts&+&-& Cheetah\\
      &Meta Policy \citep{clavera2018model}&Meta-ensembles& Short rollouts &+&-& Cheetah\\
%      Visual Foresight& Video Prediction&MPC&?&manipulation\\
      &GATS  \citep{azizzadenesheli2018surprising}& Pix2pix&MCTS &+&-& Cheetah\\
      &Policy Optim  \citep{janner2019trust}&Ensemble& Short rollouts
                                                             &+&-&Cheetah\\
      
              &Video-prediction \citep{oh2015action}&CNN/LSTM&Action&+&+&Atari\\
              &VPN \citep{oh2017value}&CNN encoder&$d$-step &+&+&Atari\\
%              &SOLAR \citep{zhang2018solar}&PGM-LQS& Local model&-&+&Reacher\\
              &SimPLe \citep{kaiser2019model}&VAE, LSTM&MPC&+&+&Atari\\
              &PlaNet \citep{hafner2018learning}&RSSM (VAE/RNN) & CEM&-&+&Cheetah\\
              &Dreamer \citep{hafner2019dream}&RSSM+CNN& Imagine&-&+&Hopper\\
              &Plan2Explore \citep{sekar2020planning}&RSSM& Planning&-&+&Hopper\\
     \hline
      End-to-End Learning&VIN \citep{tamar2016value}&CNN&Rollout in network&+&-&Mazes\\
      Planning \& Transitions &VProp \citep{nardelli2018value}& CNN&Hierarch Rollouts &+&-&Maze, nav\\
       (Sect.~\ref{sec:e2e})        &TreeQN \citep{farquhar2018treeqn}&  Tree-shape Net& Plan-functions&+&+&Box-push\\
              &Planning \citep{guez2019investigation}&CNN+LSTM&
                                                               Rollouts in network&+&-&Sokoban\\

      &I2A \citep{racaniere2017imagination}&CNN/LSTM encoder&Meta-controller&+&+&Sokoban\\
              &Predictron \citep{silver2017predictron}& $k,\gamma,\lambda$-CNN-predictr& $k$-rollout&+&+&Mazes\\
              &World Model \citep{ha2018world}&VAE & CMA-ES&+&+&Car Racing\\
      &MuZero \citep{schrittwieser2019mastering}&Latent&MCTS&-&+&Atari/Go\\
      \hline\hline\\
      
    \end{tabular}
    \caption{Overview of Deep Model-Based Reinforcement Learning Methods}\label{tab:overview}
  \end{center}
\end{table*}
%
% We will discuss the approaches in this section in this order.
The   different approaches can and have been used alone and
in combination, as we will see shortly. % follows a loosely
% chronological order, although also some cross-influence has
% occurred.  
% The flower-like picture in Figure~\ref{fig:flower} illustrates how the
% approaches are related.
Table~\ref{tab:overview} provides an overview of all approaches and
methods that we will discuss in this survey. The methods are grouped into the three
main categories that were introduced above. The use of the two
orthogonal approaches by the methods (imagination and latent models) is indicated in
Table~\ref{tab:overview} in two separate columns.
The final column provides an indication of the application that the
method is used on (such as Swimmer, Chess, and Cheetah). In the next section, Sect.~\ref{sec:bench}, these
applications will be explained in more depth.

All methods in the table will be explained in detail in the remainder
of this section (for ease of reference, we will repeat the methods of each subsection in their
own table). The sections will again mention some of the
applications on which they were tested. Please refer to the 
section on Benchmarks.
 
Model-based methods work well for low-dimensional tasks where the
transition and reward 
dynamics  are relatively simple~\citep{sutton2018introduction}. While efficient methods
such as Gaussian processes can learn these models
 quickly---with few samples---they struggle to represent  complex and
discontinuous  systems~\citep{wang2019benchmarking}. Most current
model-free methods use deep neural networks to deal with problems that have such 
complex, high-dimensional, and discontinuous characteristics, leading to
a high sample complexity.
% Deep
% neural network  models can 
% %scale to large datasets with high-dimensional inputs, and can
% represent such systems more effectively.

The main challenge that the model-based reinforcement learning
algorithms in this survey thus address is as follows.
For high-dimensional tasks the curse of dimensionality 
causes data to be sparse and variance to be high. Deep  methods tend to
overfit on small datasets, and  model-free methods
use  large data sets and  have bad sample efficiency. Model-based methods that use poor
models make poor planning predictions far into the
future~\citep{talvitie2015agnostic}.
The challenge is to learn deep, high-dimensional transition functions from limited
data, that can account for model uncertainty, and plan over these
models to achieve  policy and value functions that perform as well or
better than model-free methods.
%The challenge is to find model-based methods that have good sample
%efficiency, yet can model complex high dimensional functions well
%(using deep neural networks) making good predictions into the future.
%% While all algorithms discussed in this survey aim at achieving this
%% goal,  we will see that the algorithms of the later  approaches in
%% general come closest to this goal. 

%% We wish to note that the field of deep model-based reinforcement
%% learning is and has been very active. Recent years have seen many important
%% developments. The sheer volume of papers means that many important
%% works could not be included in this survey. Instead, our aim is necessarily more modest, to provide
%% an overview of the main 
%% ideas in recent works, showing how ideas influenced eachother.

%With this caveat out of the way,
We will now discuss the
algorithms. We will discuss
(1) methods that use explicit planning on given transitions,
(2) use explicit planning on a learned transition model, and
(3) use end-to-end learning of planning and transitions.
We will encounter the first occurrence of   hybrid imagination
%(using environment samples for updating both policy and transition models)
and latent models approaches in the second
section, on explicit planning/learned transitions.

\subsection{Explicit Planning on Given
  Transitions}\label{sec:selfplay}
% {\bf Given transitions: provided by rules of the domain. Use explicit
% planning algorithms (not learned). For high-dimensional problems, the policy and Value models are learned
% and are used in explicit planners. This approach is based on classic
% low-dimensional planning approaches of heursitic search cite
% plaat2020. A smart way of self-learning has been created, that creates
% a natural way of curriculum learning, tabula rasa.
% Conclusion: given models make efficient and very hihg performance possible.
% }
The first  approach in model-based learning is when the transition
and reward model is
provided clearly in the rules of the problem. This is the case, for example,
in games such as Go and chess. Table~\ref{tab:self} summarizes the
approaches of this subsection. Note the addition of the reinforcement
learning method in an extra column.

With this approach high performing results
have recently been achieved on large and complex domains. These
results have been achieved by combining classical,
explicit, heuristic search planning algorithms such as
Alpha-beta and MCTS~\citep{knuth1975analysis,browne2012survey,plaat2020learning},
and deep learning with self-play, achieving tabula rasa curriculum learning.
Curriculum
learning is based on the observation that a difficult problem is
learned more quickly by first learning a sequence of easy, but related
problems---just as we  teach school children easy
  concepts (such as  addition) first before we teach them harder concepts
  (such as  multiplication, or logarithms).
% In two-player games the environment of the agent can
% be played by an
% identical copy of the agent itself,. Model-based planning is 
% augmented with  self-play tournaments to generate  examples
% for training the policy and value functions.

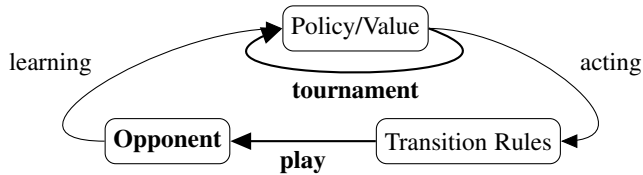
\begin{figure}[t]
  \begin{center}
    \begin{tikzpicture}[>=triangle 45,
  desc/.style={
		scale=1.0,
		rectangle,
		rounded corners,
		draw=black, 
		}]

  \node [desc,minimum height=0.6cm] (tm) at   (4,1) {Transition Rules};
    \node [desc,minimum height=0.6cm] (env) at   (0,1) {\bf Opponent};
  \draw (env.east) edge[<-,in=180,out=0,looseness=2.5,thick] node[below]
  {\bf play} (tm.west);
%  \path[desc] (game.south) edge[semithick,loop above,<-,min
%  distance=20mm,out=210,in=330,looseness=2.5] node[below] {\em game}  (game.south);

  \node [desc,minimum height=0.6cm] (pol) at   (2.5,2.5) {Policy/Value};
  \draw (env.west) edge[->,in=180,out=180,looseness=1.5] node[auto] (tour)
  {learning} (pol.west);
  
  \draw (pol.east) edge[->,in=200,out=340,looseness=3,thick] node[auto]
  {\bf tournament} (pol.west);
  
  %\path[desc] (tour.north) edge[semithick,loop left,<-,min
  %distance=15mm,in=120,out=240,looseness=1] node[right] {\ \ \ \ tournament}  (tour.north);
  
  \draw (pol.east) edge[->,out=360,in=0,looseness=1.5] node[auto] {acting} (tm.east);
  
\end{tikzpicture}
    \caption{Explicit Planning/Given Transitions}\label{fig:self}
  \end{center}
\end{figure}

In self-play the agent plays against the environment, which is also
the same agent with the same network, see Figure~\ref{fig:self}. The states and actions in the games are
then used by a deep learning 
system to improve the policy and value functions. These
functions are used as the selection and evaluation functions in MCTS,
and thus improving them improves the quality of play of MCTS.  This has the effect that as the
agent is getting smarter, so is the environment. A virtuous circle of a
mutually increasing level of play is the result, a natural form of curriculum
learning~\citep{bengio2009curriculum}.
%In self-play a planning agent such as MCTS plays a number of games against the other
%MCTS agent.
A sequence of ever-improving tournaments is played, in which a game can
be learned to play from scratch, from zero-knowledge to world champion
level~\citep{silver2017mastering}.

The concept of
self-play was invented in multiple
places and has a long history in two-agent game playing
AI. Three well-known examples are Samuel's checkers
player~\citep{samuel1959some}, Tesauro's Backgammon
player~\citep{tesauro1995td,tesauro2002programming} and DeepMind's
Alpha(Go) Zero~\citep{silver2017mastering,silver2018general}. 

% Programs for two-player games are model-based learners. In two-player
% games the transition function is known, since it is determined
% by the rules for moving the pieces of the game. The reward-function is
% also known from the rules of the game. 
% % 13 (TDGammon), 14 (AlphaZero), 15 (ExIt),

% %\subsubsection{General Approach}
% %Self-play in which the environment evolves as the transition
% %model is improved.
% The transition model is used by the agent, and also by the
% environment. A form of transfer learning occurs: the environment becomes
% smarter, as the agent becomes smarter. 

\begin{table*}[ht]
  \begin{center}
    \begin{tabular}{lllll}
      {\em Approach}&{\em Learning}&{\em Planning}& {\em Reinforcement
                                                    Learning}&{\em Application}\\ \hline\hline
      TD-Gammon \citep{tesauro1995td}&Fully connected net&Alpha-beta& Temporal Difference&Backgammon\\
      Expert Iteration \citep{anthony2017thinking}&Pol/Val CNN&MCTS &Curriculum&Hex\\
      Alpha(Go) Zero  \citep{silver2017mastering}&Pol/Val ResNet&MCTS&Curriculum&Go/chess/shogi\\
      Single Agent  \citep{feng2020solving}&ResNet&MCTS&Curriculum&Sokoban\\
      \\
    \end{tabular}
    \caption{Overview of Explicit Planning/Given Transitionds Methods}\label{tab:self}
  \end{center}
\end{table*}

Let us discuss some of the self-play approaches.

% \subsubsection{TD-Gammon}
%{\bf neural network}
%{\em Name:}
{\bf TD-Gammon}~\citep{tesauro1995td} is a Backgammon playing
program %{\em Model:}
that uses a small neural network with a single fully connected hidden layer
with just 80 hidden units %{\em Planner:}
and a small (two-level deep)
Alpha-beta search~\citep{knuth1975analysis}.
% {\em Learning:} {\em Problem:}
It teaches itself to play
Backgammon from scratch using temporal-difference learning. A small
neural network learns  the value function.  
% {\em Performance:}
TD-Gammon was the first Backgammon program to play
at World-Champion level, %{\em Rationale: link with other approach}
and
the first program to successfully use a self-learning curriculum approach in game
playing since
Samuel's checkers program~\citep{samuel1959some}. 

% \subsubsection{ExIt}
%{\bf MCTS+network}
%{\em Rationale: link with other approach}
An approach similar to the AlphaGo and  AlphaZero programs
was presented as {\bf Expert
Iteration}~\citep{anthony2017thinking}.
% {\em Problem:}
The problem was again how to learn to
play a complex game from scratch. % {\em Name:}
Expert Iteration (ExIt) combines
search-based planning (the expert) with deep learning (by
iteration).
%
% {\em Learning:}
The expert finds improvements to the
current policy. % {\em Model:}
ExIt uses a single multi-task neural network, for the
policy and the value function.  % {\em Planner:}
The planner  uses the
neural network policy and value estimates to 
improve the quality of its plans, resulting in a cycle of mutual
improvement. The planner in ExIt is MCTS.  ExIt uses a version with
rollouts. % {\em Performance:}
ExIt was used with the boardgame Hex~\citep{hayward2019hex}, and compared favorably
against a strong MCTS-only program MoHex~\citep{arneson2010monte}.
A further development of ExIt is Policy Gradient Search, which uses
planning without an explicit search tree~\citep{anthony2019policy}.

%\subsubsection{AlphaZero}
%{\bf MCTS+network for Go, chess, shogi}
%{\em Name:}
{\bf AlphaZero}, and its predecessor AlphaGo Zero, are self-play
curriculum learning programs
that were developed by a team of 
researchers~\citep{silver2018general,silver2017mastering}. 
%
%{\em Problem:}
The programs are desiged to play complex board games full of tactics and
strategy well, specifically Go, chess, and shogi, a Japanese game similar to chess, but
more complex~\citep{iida2002computer}.  %{\em Rationale: link with other approach}
AlphaZero and AlphaGo Zero are  self-play model-based reinforcement
learning programs. % {\em Learning:}
The environment against which they play is the
same program as the agent that is learning to play. The transition
function and the reward function are defined by the rules of the
game. The goal is to learn optimal policy and value functions.
%
% {\em Model:}
AlphaZero uses a single neural network, a 19-block residual network
with a value head and a policy head. For each different game---Go,
chess, shogi---it uses different
input and output layers, but the hidden layers are identical, and so
is the rest of the architecture and  the hyperparameters that govern
the learning process. The 
loss-function  is the sum of the policy-loss and the
value-loss~\citep{wang2019alternative}. 
%
%{\em Planner:}
The planning algorithm is based on Monte Carlo Tree
Search~\citep{browne2012survey,coulom2006efficient} although 
it does not perform random rollouts. Instead, it uses the value head of the
resnet for evaluation and the policy head of the ResNet to augment the
UCT selection function~\citep{kocsis2006bandit}, as in P-UCT~\citep{rosin2011multi}.
%5
%
%
\begin{figure}[t]
  \centering{
        \begin{tikzpicture}[>=triangle 45,
  desc/.style={
                scale=0.9,
		rectangle,
		rounded corners,
		draw=black,
		}]
  \node [desc] (search) at    (0,2) {{search}};
  \node [desc] (eval) at    (3,2) { {eval}};
  \node [desc] (train) at       (1.5,0.5) {{train}};

  \draw (search.south) edge[semithick,->,out=270,in=180,looseness=1] node[left] {\small examples}  (train.west);
%  \draw (train.east) edge[semithick,->,in=250,out=10, looseness=3] node {\small\em iterate/feed back new net}  (eval.south);
  
  \path (eval.south) edge[loop,semithick,<-,in=0,out=270,looseness=1]
  node[right] {\small  \ new  net}  (train.east);
  \path[desc] (search.east) edge[semithick,loop above,<-,min distance=7mm,in=160,out=20,looseness=1] node[above] {\small  reward}  (eval.west);

\end{tikzpicture}
     \caption{Self-Play/Curriculum Learning Loop}\label{fig:selfplay}}
 \end{figure}
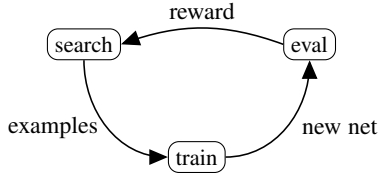
The
residual network is used in the evaluation and selection of MCTS. The self-play
mechanism starts from a randomly initialized resnet.  MCTS is used 
to play a tournament  of games, to generate training  positions for
the resnet to be trained on, using a DQN-style replay
buffer~\citep{mnih2015human}.  This trained resnet is then again used
by MCTS in the next training tournament to generate training
positions, etc., see Figure~\ref{fig:selfplay}. Self-play feeds on
itself in multiple ways, and achieving stable learning is a challenging
task, requiring judicious tuning, exploration, and much training.
%
%{\em Performance:}
AlphaZero is currently the worldwide strongest player in Go, chess,
and shogi~\citep{silver2018general}. 
 
The success of curriculum learning in two-player self-play has
inspired work on {\bf single-agent curriculum learning}. These
single-agent approaches do not do self-play, but do use curriculum learning.  Laterre et
al.\ introduce the Ranked Reward method for solving bin packing
problems~\citep{laterre2018ranked} and Wang et al.\ presented a method for
Morpion Solitaire~\citep{wang2020tackling}. Feng et al.\ 
use an AlphaZero based approach to solve hard Sokoban
instances~\citep{feng2020solving}. Their model is an 8 block standard
residual network, with MCTS as planner. Solving Sokoban instances is a
hard problem in single-agent combinatorial search. The curriculum approach,
where the agent learns to solve easy instances before it tries to
solve harder instances, is a natural fit. In two-player games, a
curriculum is generated in self-play. Feng et al. create a curriculum
in a different way, by constructing simpler subproblems from hard
instances, using the fact that Sokoban problems have a natural
hierarchical structure. As in AlphaZero, the problem learns from scratch, no Sokoban
heuristics are provided to the solver. This approach was able to solve
harder Sokoban instances than had been solved before.

\subsubsection*{Conclusion}
In self-play curriculum learning the opponent has the  same model as the agent. The
opponent is the environment of the agent. As the agent learns, so does
its opponent, providing tougher counterplay, teaching the agent
more. The agent is exposed to curriculum learning, a sequence of
increasingly harder learning tasks. In this way, learning strong play
has been achieved in Backgammon, Go, chess and 
shogi~\citep{tesauro1995temporal,silver2018general}.

In two-agent search  a natural idea is to duplicate the agent as
the environment, creating a self-play system.
Self-play has been used in planning (as minimax), with policy
learning, and in combination with latent models. % In MuZero self-play 
% was able to learn the rules of games as different as chess and Atari
% with the same abstract model-based reinforcement learning architecture.
%
Self-generated curriculum learning is a powerful paradigm.
Work is under way to see if it can be applied to single-agent
problems as
well~\citep{narvekar2020curriculum,feng2020solving,doan2019line,laterre2018ranked},
and in multi-agent (real-time strategy) games, addressing problems
with specialization of two-agent games (Sect.~\ref{sec:rts} ~\citep{vinyals2019grandmaster}).

\subsection{Explicit Planning on Learned Transitions}\label{sec:foresight}
% {\bf In the first approach the transition model was given by the
%   problem and used by
%   a classic, explicit, planning algorithm. When the model transitions
%   can  not be derived easily from  the problem domain, then the second
%   approach is to learn them using supervised machine learning
%   algorithms, using backpropagation on the environment samples. The transition model can then
%   be used with the classical planning algorithms.}

% {\bf Two variants of this approach are also discussed in this
%   subsection. That is (1) dual use of the environment samples for
%   updating the policy/value function and the transirtion model and then
%   updatinng the policy/value function by the planner, and (2) using
%   abstract models with a latent representation for the transition
%   model. (These latent models are smaller, and lower capacity models
%   need fewer samples.)}

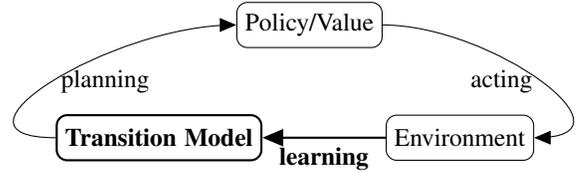
\begin{figure}[t]
  \centering
    \begin{tikzpicture}[>=triangle 45,
  desc/.style={
		scale=1.0,
		rectangle,
		rounded corners,
		draw=black, 
		}]

  \node [desc,minimum height=0.6cm] (env) at   (4.5,0.5) {Environment};
  \node [desc,minimum height=0.6cm,thick] (tm) at   (0.5,0.5) {\bf Transition Model};
  \node [desc,minimum height=0.6cm] (pol) at   (2.5,2) {Policy/Value};
  \draw (env.west) edge[->,in=0,out=180,looseness=2.5,thick] node[below]
  {\bf learning} (tm.east);
  \draw (pol.east) edge[->,out=0,in=0,looseness=1.5] node[left] {acting} (env.east);
  \draw (tm.west) edge[->,out=180,in=180,looseness=1.5] node[right]
  {planning} (pol.west);

\end{tikzpicture}
    \caption{Explicit Planning/Learned Transitions}\label{fig:learn}
  
\end{figure}

In the previous section, transition rules could be derived from the
problem directly (by inspection). In many problems, this is not the
case, and we have to resort to sampling the environment to learn a
model of the transitions. The second category of algorithms of this
survey is to learn the tranition
model  by
backpropagation from environment samples. This learned model is then still used by classical, explicit,
planning algorithms, as before.
% \subsection{Learning+ Derivation}
% During a model-free learning process information about the policy or value
% function is gathered by sampling the environment. Model-based
% reinforcement learning is about 
% using this information in a model to improve the learning process.  % There are
% two principled 
% ways in the literature in which this information can be used to
% improve the learning: by using the derivatives and by what is
% called ``imagination:'' generating extra data
% through planning.
We will  discuss various approaches where the transition model is learned with
supervised learning methods such as backpropagation through
time~\citep{werbos1988generalization}, see Figure~\ref{fig:learn}.

\begin{algorithm}[t]
  \begin{algorithmic}
    \Repeat
    \State Sample env $E$ to generate data $D=(s, a, r', s')$ 
    \State Use $D$ to learn $T_a(s,s')$
    \State Use $T$ to update policy $\pi(s, a)$ by planning 
    \Until $\pi$ converges
  \end{algorithmic}
  \caption{Explicit Planning/Learned Transitions}\label{lst:back}
\end{algorithm}
%
%
%There was model-free learning. Iw worked very well. But was sample
%inefficient. So people sought ways to improve sample efficeincy. They
%thought of derivatives, to improve the dynamics of the model.
%So derivatives are used to also update the model, in addition to samples.
%One of the earliest things that people did trying to imporve
%performance of model-free is to make extra use of the present
%derivatives in teh model.
%
% 9 (ensemble), 19 (SVG), 20 (PILCO), 21 (iLQG), 22 (GPS), 24 (local),
% 28 (PETS)
%
%
% Derivatives of model-free learned transition model to
% improve the model.
%
% Contrary to Dyna-style algorithms, where the learned dynamics models
% are used to provide imagined data, policy search with backpropagation
% through time exploits the model derivatives. Consequently, these
% algorithms are able to compute the analytic gradient of the RL
% objective with respect to the policy, and improve the policy
% accordingly.
% Updating model without imagination/simulation, but with derivatives
% of samples so far.
%
%
%
% When a differentiable environment  is present, then the policy,
% model, and reward function can be used to compute an analytic
% policy gradient by backpropagation. Several methods have been devised
% that use this approach of supervised learning by backpropagation.
%
%
% \todo{first random variables, then guided by real samples, then MPC
%   and video prediction, then uncertainy with ensembles. Improving
%   line up. }
\begin{table*}[ht]
  \begin{center}
    \begin{tabular}{llll}
      {\em Approach}&{\em Learning}&{\em Planning}&{\em Application}\\ \hline\hline
      PILCO~\citep{deisenroth2011pilco}&Gaussian Processes& Gradient based &Pendulum\\
      iLQG~\citep{tassa2012synthesis}& Quadratic Non-linear &MPC& Humanoid\\
      GPS~\citep{levine2014learning}& iLQG& Trajectory&Swimmer\\
      SVG~\citep{heess2015learning}&Value Gradients& Trajectory & Swimmer\\
      PETS \citep{chua2018deep}& Uncertainty Ensemble&MPC& Cheetah\\
%      ME-TRPO\citep{kurutach2018model}&Ensemble&Sample&Cheetah\\
      Visual Foresight \citep{finn2017deep}& Video Prediction&MPC&Manipulation\\
      \\
    \end{tabular}
    \caption{Overview of  Explicit Planning/Learned Transitions Methods}\label{tab:learn}
  \end{center}
\end{table*}

Algorithm~\ref{lst:back} shows the steps of using explicit planning
and transition learning by
backpropagation. Table~\ref{tab:learn} summarizes the approaches of
this subsection, showing both the \emph{learning} and the
\emph{planning} approach. Two variants of this approach are also discussed in this
 subsection: hybrid imagination and latent models, see
 Table~\ref{tab:imag}  and Table~\ref{tab:abstract}.
 
We will first see  how simple Gaussian Processes and quadratic
methods can create predictive transition models. Next, precision is
improved with 
trajectory methods, and we make the step to  video
prediction methods. Finally, methods that focus on uncertainty and
ensemble methods will be introduced. We know that deep neural nets
need much data and learn slowly, or 
 will overfit. Uncertainty modeling is based on
the insight that  early in the training the model has seen little data,
and tends to overfit, and later on, as it has seen more data, it may underfit.
This issue can be mitigated by incorporating uncertainty into the
dynamics models, as we shall see in the later methods~\citep{chua2018deep}. 
% , and self-supervised learning
% (variational autoencoders). 

% \todo{There should be a red line, a main thread, in this
%   part. models start small, get bigger, because of what methods? start
% gaussian, and then ensemble succeeds in biggest? need uncertainty?
% prove the power of the uncertainty}

%\todo{Problem, Name, Approach: Model/Planner/RL-alg, Rationale, Performance}

%\subsubsection{Gaussian Processes}
%{\bf Gaussian Processes}
%\todo{OK, nu is de tekst te begrijpen, maar wil ik zoveel details?}
%{\em Problem:}
For smaller models, environment samples can be used to approximate a  transition model as a
{\bf Gaussian Process} of random variables.
% {\em Name:}
This approach is followed in PILCO, which stands for Probabilistic Inference for Learning
Control, see~\citep{deisenroth2011pilco,deisenroth2013gaussian,kamthe2017data}. 
%
%Observation/Abstract: Observation
%Simulator/Imagination: use derivatives from model
%
%{\em Model:}
Gaussian Processes can
accurately learn simple processes with good sample
efficiency~\citep{bishop2006pattern}, although for high dimensional
problems they need more samples. % Other methods were tried to improve performance.
PILCO treats the transition model $T_a(s,s^\prime)$  as a probabilistic
function of the  environment samples. % The policy $\pi_\theta$
% is  trained to maximize the reward $R$. The learning process interleaves sampling (using the
% current policy) with improving the policy.
%  An analytic derivative
% is calculated of the reward relative to the policy parameters
% $\theta$.
The planner  improves the policy based on the analytic gradients
relative to the policy parameters $\theta$. 
%
%
%
%{\em Rationale: link with other approach}
%
%{\em Performance:}
%
PILCO has been used to optimize small
problems such as
Mountain car and Cartpole pendulum swings, for which it works well. Although they
achieve model learning  using higher order model information, %model inference with
Gaussian Processes do not scale to high dimensional environments,
and the method is limited to smaller applications.

%\subsubsection{Iterative Linear  Quadratic-Gaussian}
%{\bf iLQG control}

%{\em Problem:}
A related method %Gaussian Processes
uses %PILCO is the use of 
a trajectory optimization approach with nonlinear least-squares
optimization.  In control theory, the linear–quadratic–Gaussian (LQG) control problem
is one of the most fundamental optimal control problems. 
% {\em Name:}
{\bf Iterative LQG}~\citep{tassa2012synthesis}   is the control
analog of the Gauss-Newton method for nonlinear least-squares
optimization. 
% in Synthesis and Stabilization of Complex Behaviors through Online
% Trajectory Optimization. 
%{\em Rationale: link with other approach}
%{\em Model:}
In contrast to PILCO, the model learner uses quadratic approximation on the  reward
function and linear approximation of the transition function.
%
% {\em Planner:}
The planning part of this method uses a form of online trajectory optimization,
model-predictive control (MPC), in which  step-by-step real-time local
optimization  is used, as opposed to full-problem
optimization~\citep{richards2005robust}. 
% It has been applied in complex humanoid MuJoCo manipulation.
By using many further improvements throughout
the MPC pipeline, including the trajectory optimization algorithm,
the physics engine, and cost function design,
%
% {\em Performance:}
Tassa et al.\ were able to achieve
near-real-time performance in humanoid simulated robot manipulation tasks, such as
grasping. 
%
%
% The problem is
% to determine an output feedback law that is optimal in the sense of
% minimizing the expected value of a quadratic cost criterion. LQR
% assumes the model is locally linear and time-varied. 
%
% MPC is based on repeatedly solving a finite-horizon op- timal control problem.
% trajectory optimization method (
%
%
% Observation/Abstract: Observation
%showing that local trajectory optimizationn could achieve good
%results in the robotic manipulation domain.

%\subsubsection{Guided Policy Search}
%{\bf iLQG Guided policy}

%{\em Problem:}
Another trajectory optimization method takes its inspiration from model-free learning.
%Although model-free (direct) policy search methods can perform well on complex
%high-dimensional problems, it is  at the cost of high sample
%complexity.
% {\em Name:}
Levine and Koltun \citep{levine2013guided}
introduce {\bf Guided
Policy Search} (GPS) in which the search uses trajectory optimization to  avoid poor local optima. In GPS, the parameterized policy
is trained in a supervised way with samples from a trajectory
distribution. The GPS  model optimizes the
trajectory distribution  for cost and the current policy,
to create a good training set for the policy.
%
% {\em Rationale: link with other  approach}
Guiding samples are  generated by differential dynamic programming and are
incorporated into the
policy with regularized importance sampling.
%to learn neural network controllers.
%
% {\em Model:}
%Concerning the planning of  GPS methods, 
% {\em Planner:}
In contrast to the previous methods,  GPS algorithms can train
complex policies with thousands of
parameters. 
In a sense, Guided Policy Search transforms  the iLQG controller  into a
neural network policy $\pi_\theta$ with a trust region in which the new
controller does not deviate too much from the samples~\citep{levine2014learning,finn2016guided,montgomery2016guided}.
%
% {\em Performance:}
GPS has been evaluated on planar swimming,
hopping, and walking, as well as simulated 3D humanoid
running.

%\subsubsection{Stochastic Value Gradients}
%{\bf Sampling gradients}
%{\em Name:}
Another attempt at increasing the accuracy
of learned parameterized transition models in continuous control problems is 
{\bf Stochastic Value Gradients} (SVG)~\citep{heess2015learning}. 
% {\em Problem:}
%Learned transition models are often inaccurate.
%SVG introduces a method to
It
mitigates learned model inaccuracy by computing value gradients along the real environment trajectories
instead of planned ones.
%This reduces the impact of
%model error, because models are only used to compute policy gradients,
%not for prediction of actions or states.
%SVG  is introduced in.
% The policy
% is improved with the analytic gradient of the real policy
% trajectories.
%{\em Model:}
The mismatch between  predicted  and real
transitions is addressed with re-parametrization  and
backpropagation through the stochastic samples. 
%
% Methods such as PILCO that compute an analytic policy gradient by
% backpropagation of reward along a trajectory and such as xxx that
% estimate future rewards using a learned value function (a critic) and
% compute policy gradients from subsequences of trajectories. Metjhods
% that compute the policy gradient through backpropagation are referred
% to as value
% gradient methods. 
% SVG uses re-parameterization to extend the scope of value gradients
% algorithms to optimization of stochastic policies
% in stochastic environments.
%
% An environment transition model, value function, and policy
%can be learned jointly with neural networks based only on environment
%in- teraction. 
%
%Let us compare SVG to PILCO and GPS.
%{\em Rationale: link with other approach}
In comparison, PILCO  uses Gaussian process models to
compute analytic policy
gradients that are sensitive to model-uncertainty and %{\em Planner:}
GPS optimizes policies with the aid of a
stochastic trajectory optimizer and locally-linear models.  SVG in
contrast  focuses on  global neural network  value function approximators. 
%
%
%{\em Performance:}
SVG  results are reported on simulated robotics applications  in 
Swimmer, Reacher, Gripper, Monoped, Half-Cheetah, and Walker.

% \subsubsection{Probabilistic Ensembles with Trajectory Sampling}
% {\em Problem:}
Other methods also focus on uncertainty in  high dimensional modeling, but use ensembles.
%{\bf ensemble}
%Another probabilistic approach to high dimensional predictive function
%approximation uses ensembles with trajectory sampling.  {\em Name:}
Chua et
al.\ propose %a new algorithm called
{\bf probabilistic  ensembles with trajectory sampling} (PETS)~\citep{chua2018deep}.
% {\em Model:}
The learned
transition model of PETS has an uncertainty-aware deep network, which
is combined  with sampling-based uncertainty propagation.
%
% {\em Rationale: link with other approach}
% Model capacity is critical in the success of model-based
% reinforcement learning
% methods. While efficient models such as Gaussian processes can learn
% quickly, they struggle to represent high dimensional and
% discontinuous dynamical systems, that can be modeled well by
% neural networks. However,
% prior work has generally
% found that expressive parametric models, such as deep neural networks,
% generally do not produce model-based RL algorithms that are
% competitive with their model-free counterparts in terms of asymptotic
% performance [Nagabandi et al., 2017], and often even found that
% simpler time-varying linear models can outperform expressive neural
% network models [Levine et al., 2016, Gu et al., 2016].
% A major challenge is building a model that performs well in low and
% high data regimes: in the early stages of training, data is scarce,
% and highly expressive function approximators are liable to overfit;
% In the later stages of training, data is plentiful, but for systems
% with complex dynamics, simple function approximators might
% underfit. While Bayesian models such as GPs perform well in low-data
% regimes, they do not scale favorably 
% with dimensionality and often use kernels ill-suited for discontinuous
% dynamics [Calandra et al., 2016], which is typical of robots
% interacting through contacts.
 PETS uses  a combination of probabilistic
ensembles~\citep{lakshminarayanan2017simple}.  The dynamics are
modelled by an ensemble of probabilistic neural network models in a
model-predictive control setting (the agent only applies the
first action from the optimal sequence and re-plans at every
time-step)~\citep{nagabandi2018neural}.
%PETS-CEM uses the cross-entropy-methods to obtain a better solution~\citep{de2005tutorial,botev2013cross}.
%
%{\em Performance:}
Chua et al.\ report experiments on  simulated robot tasks
such as Half-Cheetah, Pusher, Reacher. Performance on these tasks is
reported to 
approach asymptotic model-free baselines, stressing the importance
of uncertainty estimation in model-baed reinforcement learning.

% %\subsubsection{Model Ensemble TRPO}
% %{\bf ensemble trpo}
% %{\em Problem:}
% Kurutach et al.\ also use ensembles to model
% uncertainty in a different way, extending {\bf ensembles with trust
%   regions}~\citep{kurutach2018model}. % also focus on modeling
% % uncertainty. {\em Name:}
% They introduce model-ensemble trust-region policy
% optimization after finding that the policy optimization stage steers the
% optimization towards areas of inaccuracy where data is scarce. In
% ME-TRPO an 
% ensemble of deep neural networks is used to  maintain model
% uncertainty, while
% TRPO~\citep{schulman2015trust} is used to counter vanishing and exploding
% gradients~\citep{hui2018model,goodfellow2016deep}. % that can be encountered in approaches based on backpropagation.
% In the planner, each imagined step is sampled from the
% ensemble predictions.
% %{\em Performance:}
% On applications from simulated robotics (Snake, Half-Cheetah, Ant, Swimmer,
% hopper) ME-TRPO achieves 100 times better sample efficiency than
% model-free baselines. 
% %
% An overview of the use of derivatives in model-based reinforcement
% learning can be found in~\citep{hui2018model}.

% \subsubsection{Visual Foresight}\label{sec:foresight}
% {\em Problem:}
An important problem in robotics is to learn arm manipulation directly
from video camera input by seeing which movements work and which
fail. The video input provides a high dimensional 
and difficult input and increases problem size and complexity substantially.
% {\bf video prediction}
%\todo{Maybe do video prediction at the end?}
%Learning to manipulate robotic arms direcly from  video camera input is an
%important goal to which  reinforcement learning can contribute. It
%also poses important challenges to overcome.
%
%{\em Name:}
Both Finn et al.\ and Ebert et al.\ report on  learning complex robotic manipulation
skills from high-dimenstional raw sensory pixel inputs in a method
called {\bf Visual Foresight}~\citep{finn2017deep,ebert2018visual}. 
% {\em Rationale: link with other approach}
The aim of Visual Foresight is to generalize deep learning methods to
never-before-seen tasks and objects. %{\em Planner:}
% The method  combines
% deep action-conditioned video prediction models with model-predictive
% control.
%
% {\em Model:}
It uses a %n unsupervised
training procedure where data is
sampled according to a probability distribution. Concurrently, a 
video prediction model is trained with the samples. This model
generates the corresponding sequence of future frames based on an
image and  a sequence of actions, as in GPS.
At test time, the least-cost sequence of actions is selected in  a model-predictive control planning
framework.
%
% {\em Performance:}
Visual Foresight is able to  perform
multi-object manipulation, pushing, picking and placing, and
cloth-folding tasks.

% \subsection{Planning and Learning}
\subsubsection*{Conclusion}
Model learning with a single network works well for low-dimensional
problems. We have seen that Gaussian Process modeling 
achieves  sample efficiency and 
generalization to good policies. For high-dimensional
problems, generalization and sample efficiency deteriorate,  more
samples  are needed and policies do not perform as 
well.
We have discussed methods for improvement by  guiding policies with real samples (GPS), limiting
the scope of predictions with 
model-predictive control, and using ensembles and uncertainty aware
neural networks to model uncertainty (PETS). % Hybrid  Although impressive achievements are
% reported, researchers have kept looking for model-based methods  that could
% achieve model-free performance in high 
% dimensional problems (see  abstract models, later on).

\subsubsection{Hybrid Model-Free/Model-Based Imagination}
In the preceding subsection, we have looked at how to use environment
samples to build a transition model. Many  methods
were covered to learn transition models  with as few samples as
possible. These methods are related to supervised learning
methods. The transition model was  then used by  a planning method to optimize
the policy or value function.

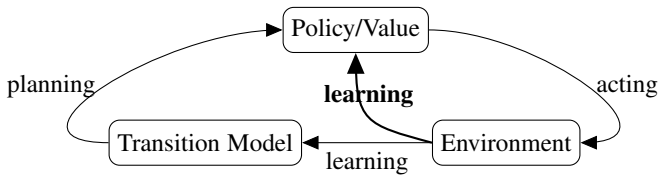
\begin{figure}[t]
  \begin{center}
    \begin{tikzpicture}[>=triangle 45,
  desc/.style={
		scale=1.0,
		rectangle,
		rounded corners,
		draw=black, 
		}]

  \node [desc,minimum height=0.6cm] (env) at   (4,0.5) {Environment};
  \node [desc,minimum height=0.6cm] (tm) at   (0,0.5) {Transition Model};
  \node [desc,minimum height=0.6cm] (pol) at   (2,2) {Policy/Value};
  \draw (env.west) edge[->,in=0,out=180,looseness=2.5] node[below]
  { learning} (tm.east);
  \draw (env.west) edge[->,in=270,out=165,looseness=1.5,thick] node[above]
  {\bf learning} (pol.south);
  \draw (pol.east) edge[->,out=0,in=0,looseness=1.5] node[right] {acting} (env.east);
  \draw (tm.west) edge[->,out=180,in=180,looseness=1.5] node[left] {planning} (pol.west);

\end{tikzpicture}
    \caption{Hybrid Model-Free/Model-Based Imagination}\label{fig:imagination}
  \end{center}
\end{figure}

We will now review methods that use a complementary approach, a hybrid model-based/model-free
approach of using the environment samples for two purposes. Here the emphasis is no longer on learning the model
but  on using it effectively. % We now focus on  generating 
% more data with the model to augment the model-free environment
% samples. The idea is to use a transition model
% to help optimize the policy by generating more data, to augment the
% sample-free data. Planning finds the optimal policy with lookahead and backup (see Section~\ref{sec:planning}). Previously the
% model-based updates were used instead of the model-free updates to the policy. In this section
% the model-based updates and model-free updates are used both to update
% the policy.
%
This approach was introduced by Sutton~\citep{sutton1990integrated,sutton1991dyna}
in the Dyna system, long before deep learning was used widely. Dyna
uses the samples to update the policy 
function directly (model-free learning) and also uses the 
samples to learn a transition model, which is then used by planning to 
augment the model-free environment-samples with the model-based
imagined ``samples.'' In this way the sample-efficiency of model-free
learning is improved quite directly.
% , with the creation of these imagined planning updates.
Figure~\ref{fig:imagination} illustrates
the working of the Dyna approach. (Note that now two arrows learn from the environment samples. Model-free learning is drawn bold.) 

Dyna was introduced for a table-based approach before deep learning
became popular. Originally,
in Dyna,  the transition model is updated directly with samples, without
learning through backpropagation, however, here we discuss only deep imagination
methods.
%Note that the policy network is now updated twice in each iteration,
%once with 
%environment samples, and once with the  planning data.
%
Algorithm~\ref{lst:dyna} shows the
steps of the algorithm (compared to  Algorithm~\ref{lst:back}, the
line in italics is new, from Algorithm~\ref{lst:free}). Note how the
policy is updated twice in each iteration, by environment sampling, and by
transition planning.

% learn the policy without direct environment
% samples for the policy. In Dyna  the model-free lookahead planning is
% called ``imagination.'' In contrast to the previous section, in Dyna
% the same method is used for optimizing the policy
% from real experience as from imagined experience, generated by planning.

\begin{table*}[h]
  \begin{center}
    \begin{tabular}{lllll}
      {\em Approach}&{\em Learning}&{\em Planning}& {\em Reinforcement
                                                    Learning}&{\em Application}\\ \hline\hline
      Local Model~\citep{gu2016continuous}& Quadratic Non-linear& Short rollouts &Q-learning& Cheetah\\
      MVE  \citep{feinberg2018model}  & Samples& Short rollouts& Actor-critic&Cheetah\\
      Meta Policy \citep{clavera2018model}&Meta-ensembles& Short rollouts & Policy optimization & Cheetah\\
%      Visual Foresight& Video Prediction&MPC&?&manipulation\\
      GATS  \citep{azizzadenesheli2018surprising}& Pix2pix&MCTS&Deep Q Network & Cheetah\\
      Policy Optim  \citep{janner2019trust}&Ensemble& Short rollouts &Soft-Actor-Critic&Cheetah\\
      Video predict  \citep{oh2015action}&CNN/LSTM&Action&Curriculum&Atari\\
      VPN  \citep{oh2017value}&CNN encoder&$d$-step &$k$-step&Mazes, Atari\\

      SimPLe  \citep{kaiser2019model}&VAE, LSTM&MPC&PPO&Atari\\
      \\
    \end{tabular}
    \caption{Overview of Hybrid Model-Free/Model-based Imagination Methods}\label{tab:imag}
  \end{center}
\end{table*}

\begin{algorithm}

  \begin{algorithmic}
    \Repeat
    \State Sample env $E$ to generate data $D=(s, a, r', s')$ 
    \State {\em Use $D$ to update policy $\pi(s, a)$}
    \State Use $D$ to learn $T_a(s,s')$
    \State Use $T$ to update policy $\pi(s, a)$ by planning 
    \Until $\pi$ converges
    \end{algorithmic}
    \caption{Hybrid Model-Free/Model-Based Imagination}\label{lst:dyna}

\end{algorithm}

We will  describe five  deep learning approaches that use
imagination to augment the sample-free data.
Table~\ref{tab:imag} summarizes these  approaches. Note that in the
next subsection more methods are  described that also use hybrid
model-free/model-based updating of the policy function. These are also
listed in Table~\ref{tab:imag}.  
%
% % \subsection{Planning + Learning, Observation, Derivatives, CNN}
% Then they thought back to the derivatives idea, and merged that.
%
% 4 (Policy rollouts), 
%
We will first see  how quadratic methods are used with imagination
rollouts. Next, short rollouts are introduced, and ensembles, to
improve the precision of the predictive model. Imagination is a hybrid
model-free/model-based approach, we will  see methods that build on successful
model-free deep learning approaches such as meta-learning and generative-adversarial
networks. 

Let us have a look at how  Dyna-style imagination works in deep model-based algorithms. 
%
%
%
%
%
% % \subsection{Planning + Learning, Observation, Distributional, CNN}
% Then they realized that point values for expcted value is a bit
% narrow, and added probability distributions, which aslo works in
% model-free.
%
% 10, 11 (Meta), 
% %\subsection{Planning + Learning, Observation, Expected, RNN}
%
%
%
%\subsubsection{Local Model}
% Continuous Deep Q-learning with Model-based
% Acceleration.
% {\bf locally linear}
Earlier, we saw that linear–quadratic–Gaussian methods were used to
improve model learning.
% {\em Problem:}
Gu et al.\ merge the backpropagation iLQG aproaches with Dyna-style
synthetic policy rollouts~\citep{gu2016continuous}.
%\todo{This is a nice first algo, since it
% builds on earlier algos}
To accelerate model-free
continuous Q-learning they combine {\bf locally linear models} with
local on-policy imagination rollouts.
%
% {\em Name:}
% Iteratively fitting local linear
% models to the latest batch of rollouts %{\em Rationale: link with other approach}
% provides sufficient local accuracy to achieve substantial improvement
% using short imagination rollouts in the vicinity of the real-world
% samples, especially on-policy rollouts.
%
%{\em Model:}
The paper introduces a version of continuous Q-learning called normalized
advantage functions,  accelerating the learning with imagination
rollouts.
%NAF provides  advantages over
%actor-critic model-free methods in continuous domains. % {\em Planner:}
Data
efficiency is improved %substantially
with model-guided exploration
using off-policy iLQG rollouts. 
%
%{\em Sampling:}
%The overall reinforcement learning algorithm that is
%used is continuous Q-learning, augmented by imagined rollouts.
%
%{\em Performance:}
As application the approach has been tested on simulated robotics tasks such as Gripper, Half-Cheetah and Reacher.

%\subsubsection{Model-Based Value Expansion}
%{\bf short rollout}
%{\em Problem:}
% An advantage of adding Dyna-style   data based on the learned model
% is that it  reduces sample complexity; a disadvantage is that
% inaccuracy in the model may be 
% ``enlarged'' by the rollouts, reducing accuracy. With imagination
% extra care must be taken to reduce bias 
% and variance. However, control of uncertainty in such models is
% challenging. 
%{\em Name:}
Feinberg et al.\ present {\bf model-based value
expansion} (MVE) which, like the previous algorithm~\citep{gu2016continuous},
controls for uncertainty 
in the deep model by only allowing imagination to fixed
depth~\citep{feinberg2018model}. %{\em Rationale: link with other approach}
% Feinberg's  approach
% is close to a deep learning version of Dyna.
%By enabling wider use of learned dynamics models within a model-free
%reinforcement learning algorithm, they improve value estimation, which,
%in turn, reduces the sample complexity of learning. 
%
% MVE is a hybrid algorithm that
% uses planning with a dynamics model
% to imagine the short-term
% horizon, in combination with regular model-free Q-learning
% to estimate the long-term
% value beyond the simulation horizon.
% {\em Model:}
% MVE improves upon model-free learning by providing higher-quality
% target values for training.
Value estimates 
are split into a near-future model-based
component and a distant future model-free component. In
contrast to stochastic value gradients (SVG), MVE works without
differentiable dynamics, which  is important since transitions can include
non-differentiable  contact
interactions~\citep{heess2015learning}.
%
% {\em Planner:}
The planning part of MVE uses short rollouts.
%
% {\em Sampling:}
The overall reinforcement learning algorithm that is
used is a combined value-policy actor-critic
setting~\citep{sutton2018introduction} and deep deterministic policy 
gradients (DDPG)~\citep{lillicrap2015continuous}.
%
%
% {\em Performance:}
As application re-implementations of simulated robotics were used such as for Cheetah,
Swimmer and Walker.
% Imagination-augmented agents (I2A) offloads all uncertainty estimation
% and model use into an implicit neu- ral network training process,
% inheriting the inefficiency of model-free methods (Racaniere et al.,
% 2017)

An ensemble approach has been used  in combination with
gradient-based meta-learning by 
%\subsubsection{Meta-Policy Optimization}
%{\bf Meta ensemble}
%{\em Name:}
Clavera et al.\ who introduced 
Model-based Reinforcement Learning via {\bf Meta-Policy
Optimization} (MP-MPO)~\citep{clavera2018model}.
%MB-MPO uses ensembles. %{\em Problem:} {\em Rationale: link with other
% approach}
This method learns an ensemble of
dynamics models and then it  learns %a policy that
%can adapt in one step  to any model in the ensemble.
%
% MB-MPO meta-learns a
% policy that can quickly adapt to any model in the ensemble with one
%5 policy gradient step.
%
%
%{\em Sampling:}
%Using the models as learned simulators, MB-MPO learns
a policy that
can be adapted quickly to any of the fitted dynamics models with one
gradient step (the MAML-like meta-learning step~\citep{finn2017model}).
% This optimization objective steers the meta-policy
% towards internalizing the parts of the dynamics prediction that are
% consistent among the ensemble while shifting the burden of behaving
% optimally with respect to differences between models towards the adaptation
% step.
%
% {\em Model:}
MB-MPO  frames model-based reinforcement learning as meta-learning a policy on a distribution
of dynamic models, in the form of an ensemble of the real environment dynamics.
The approach builds on the gradient-based meta-learning framework
MAML~\citep{finn2017model}.
% , which trains a
% parametric policy $\pi_\theta(a|s)$ to 
% quickly improve its performance on a new task with one or a few
% vanilla policy gradient steps~\citep{peters2006policy} in reinforcement
% learning for the adaptation step.
%
% {\em Planner:}
The planning part of the algorithm samples
imagined trajectories.
%
% MAML attempts to learn an initialization $x$ such that for any task
% $M_k$ (M) the policy attains 
% maximum performance in the respective task after one policy gradient
% step.
%
% Regularization effect during training. Optimizing the policy to adapt within one policy gradient step to any of the fitted models imposes a regularizing effect on the policy learning
%
% Since we sample real-environment trajec- tories using the different
% policies $\{1 , ..., K \}$ obtained by adaptation to each model, the
% collected training data is more diverse which promotes robustness of
% the dynamic models. As a result, we collect real-world data in regions where the dynamic models insufficiently approxi- mate the true dynamics. This effect accelerates correcting the imprecision of the models leading to faster improvement.
%
%
%
 % learns an ensemble of dynamics models and meta-optimizes a policy
 % for adaptation in each of the learned models. Our experimental
 % results demonstrate that meta-learning a policy over an ensemble of
 % learned models provides the recipe for reaching the same level of
 % per- formance as state-of-the-art model-free methods with
 % substantially lower sample complexity.  
%
%{\em Performance:}
MB-MPO is evaluated on continuous control benchmark
tasks in a robotics simulator: Ant, Half-Cheetah, Hopper, Swimmer, Walker.
The results reported indicate that meta-learning a policy over an
ensemble of learned models approaches  the level of performance of
model-free methods with substantially better sample complexity.

%\subsubsection{Generative Adversarial Tree Search}
%{\bf pix2pix}
%
%{\em Problem:} {\em Name:}
Another attempt to improve the accuracy and efficiency of dynamics models has been through
generative adversarial
networks~\citep{goodfellow2014generative}.
% \todo{Replace this by  Nagabandi. Or a third one?}
Azizzadenesheli et
al.\ aim to combine successes of
{\bf generative 
adversarial networks}  with planning robot motion in model-based
reinforcement learning~\citep{azizzadenesheli2018surprising}. Manipulating robot arms based on video input
is an important application in AI (see also Visual Foresight in
Section~\ref{sec:foresight},  and  the SimPLe approach, in Section~\ref{sec:simple}). % {\em Model:}
A generative dynamics model is
introduced to 
model the transition dynamics  based on the pix2pix
architecture~\citep{isola2017image}. % {\em Planner:}
For planning Monte Carlo Tree
Search~\citep{coulom2006efficient,browne2012survey}
is used. % {\em Sampling:}
%GATS uses a combination of model-free (DQN) and model-based
%learning.
% {\em Performance:}
GATS is
evaluated on Atari games such as Pong, and does not perform better
than model-free DQN~\citep{mnih2015human}.
%
% {\em Rationale: link with other approach}
% Nagabandi et al.\ use a model-based neural network
% to pre-train a model-free neural network, and achieve better overall results~\citep{nagabandi2018neural}.
% This hybrid algorithm achieves sample-efficiency gains of about 3 to 5
% on benchmarks such as swimmer, cheetah.

%\subsubsection{Policy Optimization}
% {\bf short rollouts, ensemble}

% \todo{Maybe put this algorithm later? so that it logically follow chua?. Before meta ensembles?} 
% {\em Problem:}
Achieving a good performing high-dimensional predictive model remains a challenge.
% {\em Name:}
Janner et al.\  propose in
Model-based Policy Optimization (MBPO) a new approach to {\bf short rollouts
  with ensembles}~\citep{janner2019trust}. In
this approach the  model horizon is much shorter
than  the task horizon. % {\em Planner:}
These model rollouts are combined with real
samples, and matched with plausible environment
observations~\citep{kalweit2017uncertainty}.
% {\em Model:}
MBPO uses an ensemble of probabilistic networks, as in
PETS~\citep{chua2018deep}.
%
% and ME-TRPO~\citep{kurutach2018model}. 
%
% {\em   Sampling:}
Soft-actor-critic~\citep{haarnoja2018soft}
is used as reinforcement learning method.
% {\em  Rationale: link with other approach}
Experiments show that the  policy optimization algorithm 
learns substantially faster with short rollouts than other algorithms,
while retaining asymptotic performance relative to model-free
algorithms.
%
% {\em Performance:}
The applications used are simulated robotics tasks:
Hopper, Walker, Half-Cheetah, Ant.
The method surpasses the sample
efficiency of prior model-based algorithms and matches the performance of
model-free algorithms.

\subsubsection*{Conclusion}
The hybrid imagination methods  aim to combine the advantage of model-free methods with
model-based methods in a hybrid approach augmenting ``real'' with
``imagined'' samples, to improve  sample effciency of deep model-free
learning. A problem is that
inaccuracies in the model may be enlarged in the planned rollouts.  Most methods limited lookahead
to local lookahead.  We have discussed interesting approaches combining
meta-learning and generative-adversarial networks,  and ensemble
methods learning robotic movement directly from  images. 

\subsubsection{Latent Models}\label{sec:abstract}

%\subsection{Planning + Learning, Abstract, Expected, VAE/RNN}
The next group of methods that we describe are the latent or abstract
model algorithms. Latent models are born out of the need for
more accurate predictive deep models. Latent models replace the
single transition model with separate, smaller, specialized, representation models,
for the different functions in a reinforcement learning algorithm. All
of the elements of the MDP-tuple  may now get their own
model. Planning occurs in latent space.

\begin{table*}[h]
  \begin{center}
    \begin{tabular}{lllll}
      {\em Approach}&{\em Learning}&{\em Planning}& {\em Reinforcement
                                                    Learning}&{\em Application}\\ \hline\hline
      Video predict  \citep{oh2015action}&CNN/LSTM&Action&Curriculum&Atari\\
      VPN  \citep{oh2017value}&CNN encoder&$d$-step &$k$-step&Mazes, Atari\\
%      SOLAR \citep{zhang2018solar}&PGM-LQS& Local model & LQR-FLM & Reacher\\
      SimPLe  \citep{kaiser2019model}&VAE, LSTM&MPC&PPO&Atari\\
      PlaNet \citep{hafner2018learning}&RSSM (VAE/RNN) & CEM & MPC & Cheetah\\
      Dreamer \citep{hafner2019dream}&RSSM+CNN& Imagine & Actor-Critic & Control\\
      Plan2Explore \citep{sekar2020planning}&RSSM& Planning & Few-shot & Control\\
      \\
    \end{tabular}
    \caption{Overview of Latent Modeling Methods}\label{tab:abstract}
  \end{center}
\end{table*}

Traditional deep learning  models represent
input states directly in a single model: the layers of neurons and filters are all
related in some way to the input and output of the domain, be it an
image, a sound, a text or a joystick action or arm movement. All of
the MDP functions, state, value, action, reward, policy, and discount, act on
this single model. Latent models, on the other hand, are not connected
directly to the input and output, but are connected to other
models and signals. They do not work on direct representations, but on
latent, more compact, representations. The interactions are captured in three to four
different models, such as observation, representation, transition, and
reward models. These may be smaller, lower capacity, models. They may be trained with
unsupervised or self-supervised deep learning such as variational autoencoders~\citep{kingma2013auto,kingma2019introduction} or
generative adversarial networks~\citep{goodfellow2014generative}, or recurrent networks.
Latent models use multiple
specialized  networks, one for each function to be
approximated.
The intuition
behind the use of latent models is dimension reduction:  they can better specialize
and thus have more precise predictions, or can better capture the 
essence of higher level reasoning in the input domains, and need fewer
samples (without overfitting) due to their lower capacity.

\begin{figure}[t]
  \begin{center}
    \begin{tikzpicture}[>=triangle 45,
  desc/.style={
		scale=1.0,
		rectangle,
		rounded corners,
		draw=black, 
		}]

  \node [desc,minimum height=0.6cm] (env) at   (5.5,0.5) {Environment};
  \node [desc,minimum width=3.3cm,minimum height=0.6cm,thick] (tm1) at
  (0,2.3) {\bf Transition Model};
  \node [desc,minimum width=3.3cm,minimum height=0.6cm,thick] (tm2) at
  (0,0.5) {\bf Representation Model};
  \node [desc,minimum width=3.3cm,minimum height=0.6cm,thick] (tm3) at
  (0,1.1) {\bf Observation Model};
  \node [desc,minimum width=3.3cm,minimum height=0.6cm,thick] (tm4) at
  (0,1.7) {\bf Reward Model};
  \node [desc,minimum height=0.6cm] (pol) at   (3,4) {Policy \& Value};
  \draw (env.west) edge[->,in=0,out=180,looseness=2] node[below]
  {learning} (tm2.east);
  \draw (pol.east) edge[->,out=0,in=90,looseness=1] node[right] {acting} (env.north);
  \draw (tm1.north) edge[->,out=90,in=180,looseness=1] node[left] {planning} (pol.west);

\end{tikzpicture}
    \caption{Latent Models}\label{fig:abstract}
  \end{center}
\end{figure}
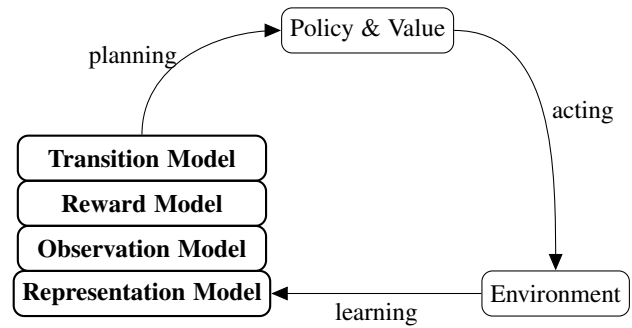

Figure~\ref{fig:abstract} illustrates the abstract (latent) learning
process (using the modules of 
Dreamer~\citep{hafner2019dream}  as an example).
%
%\todo{present the modular models better in the table. not
%  CNN+ctrl+... . And: separate fucntion of modules (repr, reward) and technologies
%  (CNN, VAE)}
%
%
%% \begin{table}[h]
%%   \begin{center}
%%     \begin{tabular}{lllll}
%%       {\em Approach}&{\em Model}&{\em Planning}& {\em Learning}&{\em Application}\\ \hline\hline
%%       I2A&CNN+ctrl+Env+LSTM&meta-controller rollout& A3C&Sokoban\\
%%       VPN&CNN+trans+outc+FCN&rollout &Imagination&Atari\\
%%       Predictron& repr+next+reward+discount& $k$-rollout&Imagination&Mazes\\
%%       PlaNet&trans+obs+enc+reward& recurrent & MPC & Cheetah\\
%%       World Models&vision+memory+control& Rollout & Imagination & Car Racing\\
%%       SOLAR&repr+trans+reward& local model & LQR-FLM & Reacher\\
%%       Dreamer&repr+obs+trans+reward& imagine & Actor-Critic & DeepMind\\
%%       Plan2Explore&CNN+trans+reward+decoder& Planning & Transfer/zero-shot & DeepMind\\
%%     \end{tabular}
%%     \caption{Overview of Abstract Modeling Approaches}\label{tab:abstract-org}
%%   \end{center}
%% \end{table}
%
%
Table~\ref{tab:abstract} summarizes the methods of this subsection
(three are also mentioned in Table~\ref{tab:imag}). 
Quite a few different latent (abstract) model approaches have been published.
Latent models work well, both for games and robotics.
Different rollout methods are proposed, such as
local rollouts, and differentiable imagination, and end-to-end
model learning and planning. Finally, latent models are applied to
transfer learning in few-shot learning. In the next subsection more
methods are  described that also use latent models (see
Table~\ref{tab:netw}  and the overview
Table~\ref{tab:overview}).  

Let us now have a look at the latent model approaches.

An important application in games and robotics  is the
long range  prediction of video images. Building a {\bf generative model
for video data} is a challenging problem involving high-dimensional natural-scene
data with temporal dynamics, introduced
by~\citep{schmidhuber1991learningfovea}. In many applications
next-frame 
prediction also depends on control or action variables, especially in
games. A first paper by~\citep{oh2015action} builds a model to predict Atari
games using a high-dimensional video  encoding model and
action-conditional transformation. The authors
describe three-step experiments with a convolutional  and with a recurrent
(LSTM) encoder. The next step performs action-conditional encoding,
after which convolutional decoding takes place. To reduce the effect
of  small
prediction errors compounding through time, a multi-step prediction
target is used. Short-term future
frames are predicted and fine-tuned to predict longer-term future frames after the
previous phase converges, using a curriculum that stabilizes
training~\citep{bengio2009curriculum}. Oh et al.\ perform
planning on an abstract, encoded, representation; showing the benefit
of acting in latent space.
Experimental results on Atari games showed generation of
visually-realistic frames  useful for 
control over up to 100-step action-conditional predictions in some
games. This architecture was developed further into the VPN approach,
which we will describe next.

%\todo{is their a logical order of development of all the abstract
%  model aproaches?}
%\todo{add little tables of the modules and  ther implementations?}

% \subsubsection{Value Prediction Network}

%{\bf latent value preediction. encoding, transition, outcome,
%  value. atari}
%  {\em Name:}
The {\bf Value Prediction
  Network} (VPN) approach~\citep{oh2017value} % {\em Problem:}
integrates model-free and
model-based reinforcement learning into a single abstract neural
network that 
consists of four modules.
%A VPN
%learns a transition model whose abstract states are trained to make
%option-conditional predictions of future values (and not of future
%observations, as a conventional network architecture would do).
%{\em Rationale: link with other approach}
For training, VPN combines temporal-difference
search~\citep{silver2012temporal} and $n$-step Q-learning~\citep{mnih2016asynchronous}. VPN
% learns to predict values via Q-learning, and rewards via supervised
% learning.
performs lookahead planning to choose actions. Classical model-based
reinforcement learning predicts future observations $T_a(s,s')$. VPN
plans future values without having to predict future observations, 
using abstract representations instead. 
%
%{\em Model:}
The VPN network architecture  consists of the modules: encoding,
transition, outcome, and value. 
%
%{\em Planner:}
The encoding module is applied to the environment observation to
produce a latent state $s$. The value, outcome, and 
transition modules work in latent space, and are recursively applied to expand the
tree.\footnote{VPN uses a convolutional neural network as the encoding module. The transition module consists
of an option-conditional convolution layer (see~\citep{oh2015action}). A residual connection from
the previous abstract-state to the next asbtract-state is
used~\citep{he2016deep}.
The outcome module is similar to the
transition module. The value module
consists of two fully-connected layers. The number of layers and
hidden units varies depending on the application domain.}
%
% VPN simulates the future based on abstract states.
It does not use
MCTS, but a simpler rollout algorithm that performs planning up to a
planning horizon.
%
% {\em Learning:}
VPN uses imagination to update the policy.
%
% {\em Performance:}
It outperforms model-free DQN on Mazes and Atari games such as Seaquest, QBert,
Krull, and Crazy Climber.
Value Prediction Networks are related to Value Iteration Networks and to the Predictron,
which we will describe next.

For robotics and games, video prediction methods are important.\label{sec:simple}
%The goal of SimPLe is to
%achieve high performance in Atari games:
Simulated policy learning, or SimPLe, 
uses {\bf stochastic video prediction} techniques~\citep{kaiser2019model}.
% {\em Rationale: link with   other approach}
SimPLe uses video frame prediction as a basis for model-based
reinforcement learning.
In contrast to Visual Foresight, SimPLe builds on
model-free work on video
prediction using variational autoencoders, recurrent world models and generative models~\citep{oh2015action,chiappa2017recurrent,leibfried2016deep} and model-based
work~\citep{oh2017value,ha2018world,azizzadenesheli2018surprising}. 
The latent model is formed with a
variational autoencoder that is used to deal with the limited horizon of
past observation
frames~\citep{babaeizadeh2017stochastic,bengio2015scheduled}. 
%to achieve the same level of
%performance with model-based learning as model-free methods
%achieve.
The model-free PPO algorithm~\citep{schulman2017proximal} is used for
policy optimization.
%SimPLe utilizes stochastic video prediction techniques and trains a policy to
%play the game within the learned model.
%{\em Performance:}
In an experimental evaluation, SimPLe is more sample efficient than the
Rainbow algorithm~\citep{hessel2017rainbow} on 26 ALE games to learn Atari
games with 100,000 sample steps (400k frames).

%  {\em Problem:}
Learning dynamics models that are accurate enough for planning is a long standing challenge, especially in image-based domains.
% {\em Name:}
PlaNet trains a model-based agent to learn the environment dynamics
from images and choose actions through planning in latent space with
both deterministic and stochastic transition elements.
PlaNet is  introduced in {\bf Planning from
  Pixels}~\citep{hafner2018learning}. 
%
%{\em Rationale: link with other approach}
%The latent dynamics  model is a recurrent state space model.
%{\em Model:}
PlaNet uses a Recurrent State Space Model (RSSM) that consists of a transition model, an observation model, a variational
encoder and a
reward model. %{\em Planner:}
Based on these models a Model-Predictive Control agent
is used to adapt its plan, replanning each step. For planning, the RSSM is
used by the Cross-Entropy-Method (CEM) to search for the best action sequence~\citep{karl2016deep,buesing2018learning,doerr2018probabilistic}. 
% {\em Learning:}
In contrast to many
model-free reinforcement learning approaches, no explicit policy or value
network is used.
%PlaNet uses model-predictivae control (MPC) to allow the agent to adapt its plan based on new observations.
%
%{\em Performance:}
PlaNet is tested on tasks from  MuJoCo and the DeepMind control suite:  Swing-up,
Reacher, Cheetah, Cup Catch. It reaches performance that is close to
strong model-free algorithms.

%\subsubsection{Dreamer}
%{\bf backprop imagination to latent models. representation,
%  observation, transition, reward. reacher }
%  {\em Name:}
A year after the  PlaNet paper was pubished \citep{hafner2019dream}
published Dream to Control: Learning Behaviors by {\bf Latent
Imagination}.
%
% {\em Problem:}
% Dreamer solves long-horizon tasks by latent imagination.
%In this behaviors are learned by backpropagating analytic gradients of learned state
%values through trajectories imagined in the state space. %of a learned
% world model.
%
% {\em Rationale: link with other approach}
World models enable interpolating between past
experience,
%{\em Model:}
% Dreamer solves long-horizon tasks by latent imagination: learn
% behaviors by backpropagating gradients of learned state values through
% trajectories imagined in compact learned wold model.
%
% {\em Learning:}
and latent models predict both actions and values.  
The latent models in Dreamer consist of a representation model, an observation
model, a transition model, and
a reward model. It allows the agent to plan (imagine) the outcomes of
potential action sequences without executing them in the
environment. It uses an actor-critic approach to learn behaviors that
consider rewards beyond the horizon.  Dreamer
backpropagates through the value model, similar to
DDPG~\citep{lillicrap2015continuous} and Soft-actor-critic~\citep{haarnoja2018soft}. 
%
% {\em Planner:}
% Dreamer solves long-horizon tasks by latent imagination. It learns
% behaviors by backpropagating gradients of learned state values through
% trajectories imagined in compact learned wold model.
% 
% {\em Performance:}
Dreamer is tested with applications from the   DeepMind control suite: 20
visual control tasks such as 
Cup, Acrobot, Hopper, Walker, Quadruped, on which it achieves good performance.

%\subsubsection{Plan2Explore}
%{\bf few-shot latent}
%
%{\em Problem:} {\em Name:}
Finally, Plan2Explore~\citep{sekar2020planning} studies how reinforcement
learning with latent models can 
be used for transfer learning, in partiular, few-shot and {\bf zero-shot
learning}~\citep{xian2017zero}. 
%
% {\em Rationale: link with other approach}
Plan2Explore is a self-supervised reinforcement learning method that
learns a world model of its environment through unsupervised
exploration, which it then  uses  to solve zero-shot and few-shot
tasks. Plan2Explore was built on PlaNet~\citep{hafner2018learning} and
Dreamer~\citep{hafner2019dream} learning dynamics models from
images, using the same latent models: image encoder (convolutional
neural network), dynamics
(recurrent state space model), reward
predictor, image decoder.
%
%{\em Model:}
With this world model, behaviors must be derived for the learning
tasks.
%Plan2Explore uses Dreamer to learn a parametric policy inside the world model.
% {\em  Planner:}
The agent
first uses planning to explore to learn a world model in a
self-supervised manner.
% {\em Learning:}
After exploration, it receives reward
functions to adapt to multiple tasks such as standing, walking,
running and flipping.
%
% {\em Performance:}
Plan2Explore achieved good zero-shot performance on the
DeepMind Control Suite (Swingup, Hopper, Pendulum, Reacher, Cup Catch,
Walker) in the sense that the agent's self-supervised zero-shot performance was
competitive to Dreamer's supervised reinforcement learning
performance.

\subsubsection*{Conclusion}
In the preceding methods we have seen  how a single network model can
be specialized in three or 
four separate models. Different rollout methods were proposed, such as
local rollouts, and differentiable imagination.
Latent, or abstract, models are a direct descendent of model learning  networks, with
different  models for
different aspects of the reinforcement learning algorithms. The latent
representations have lower capacity, allowing for greater accuracy,
better generalization and reduced sample complexity.  The smaller latent representation models are often
learned unsupervised or self-supervised,
using variational autoencoders or recurrent LSTMs.
Latent models were applied to
transfer learning in few-shot learning.

\subsection{End-to-end Learning of Planning and Transitions}\label{sec:e2e}
In the previous subsection  the  approach is (1) to learn a  transition model through
backpropagation and then
(2) to do conventional lookahead rollouts using a planning algorithm such
as value iteration, depth-limited search, or MCTS. A larger trend in
machine learning is to replace conventional algorithms by 
differentiable or gradient style approaches, that are self-learning
and self-adapting. Would it be  
possible to make the conventional 
rollouts differentiable as well? If updates can be made
differentiable, why not 
planning?

The final approach of this survey is indeed to learn both the transition model and
  planning steps end-to-end. This means that the neural
  network represents both the transition model and executes the planning
  steps with it. This is a challenge that has to do with a single neural network, but
  we will see that 
  abstract models,  with latent representations, can more
  easily be used to achieve the execution of planning steps. 

  When we
  look at the action that a neural network normally performs as a
  transformation and filter activity (selection, or classification) then it is easy to
  see than planning, which consists of state unrolling and selection,
  is not so far from what a neural network is normally used for. Note
  that especially recurrent neural networks and LSTM contain implicit state, making their
  use as a planner even easier. 

%%  The most basic difference in model-based
%% reinforcement learning between planning and learning is that planning uses the
%% transition model to find the next state, where model-free learning builds a
%% model of the policy function directly.\footnote{A second difference
%%   is, of course, that in planning the model is used to look ahead
%%   multiple steps, whereas  model-free learning is focussed on single
%%   steps.}  Can the use of the model be 
%% learned?
%, and can the full deep model-based algorithm be made
%differentiable end-to-end?

\begin{table*}[h]
  \begin{center}
    \begin{tabular}{lllll}
      {\em Approach}&{\em Learning}&{\em Planning}& {\em Reinforcement
                                                    Learning}&{\em Application}\\ \hline\hline
      VIN \citep{tamar2016value}&CNN&Rollout in network&Value Iteration&Mazes\\
      VProp \citep{nardelli2018value}& CNN&Hierarch Rollouts &Value Iteration&Navigation\\
      TreeQN \citep{farquhar2018treeqn}& Tree-shape Net& Plan-functions&DQN/Actor-Critic&Box pushing\\
      ConvLSTM \citep{guez2019investigation}&CNN+LSTM& Rollouts in network & A3C & Sokoban\\
      I2A  \citep{racaniere2017imagination}&CNN/LSTM encoder&Meta-controller& A3C&Sokoban\\
      Predictron  \citep{silver2017predictron}& $k,\gamma,\lambda$-CNN-predictr& $k$-rollout& $\lambda$-accum&Mazes\\
      World Model \citep{ha2018world}& VAE & CMA-ES & MDN-RNN & Car Racing\\

            MuZero \citep{schrittwieser2019mastering}& Latent&MCTS &Curriculum&Go/chess/shogi+Atari\\

      \\
    \end{tabular}
    \caption{Overview of End-to-End Planning/Transition Methods}\label{tab:netw}
  \end{center}
\end{table*}

\begin{figure}[t]
  \begin{center}
    \begin{tikzpicture}[>=triangle 45,
  desc/.style={
		scale=1.0,
		rectangle,
		rounded corners,
		draw=black, 
		}]

  \node [desc,minimum height=0.6cm] (env) at   (4,0.5) {Environment};
  \node [desc,minimum height=0.6cm] (tm) at   (0,0.5) {(Latent) Trans \bf Network};
  \node [desc,minimum height=0.6cm] (pol) at   (2,2) {Policy/Value};
  \draw (env.west) edge[->,in=0,out=180,looseness=2.5,thick] node[below]
  {learning} (tm.east);
  \draw (pol.east) edge[->,out=0,in=0,looseness=1.5,thick] node[left] {acting} (env.east);
  \draw (tm.west) edge[->,out=180,in=180,looseness=1.5,thick] node
  {{\bf differentiable} planning} (pol.west);

\end{tikzpicture}
    \caption{End-to-End Planning/Transitions}\label{fig:network}
  \end{center}
\end{figure}
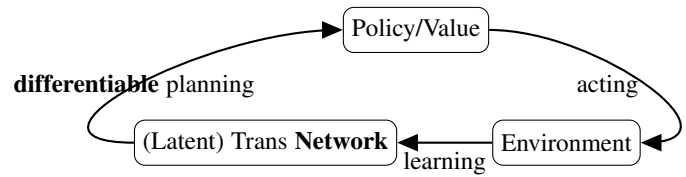

Some progress has been made with this idea. %% . We will  survey
%% approaches to integrate the planning step into the differentiable
%% learning step. Several ideas have been used
One approach is to map
the planning iterations onto the layers of a deep neural network, with
each layer representing a lookahead step. The transition \emph{model} becomes
embedded in a transition \emph{network}, see Figure~\ref{fig:network}.

In this way, the planner becomes part of one large trained  end-to-end
agent. (In the figure 
the full circle is made bold to signal end-to-end learning.)
Table~\ref{tab:netw} summarizes the approaches of this subsection.
We will see how the iterations of value iteration can be implemented
in the layers of a convolutional neural network (CNN).
Next, two variations of this method are presented, and a
way to implement planning with convolutional LSTM modules. All these
approaches implement differentiable, trainable, planning algorithms,
that can generalize to different inputs. The later methods use
elaborate schemes with
latent models so that the learning can be applied to different application domains.

% In a more general sense, planing and learning can be combined in more
% ways than backpropagation and look ahead. Planningn algoritihms can be
% put inside a neuron and the whole network can function as a plannning
% algorithm. We will look at different approaches.

% %\subsection{Planning + Learning, Observation, Expected, Planning in Network}
% There were very creative people who thought of replacing the neurons
% by MCTS nodes, and also people who realized that there is a
% correspondence between look-ahead and deep-layer progression, and do
% MCTS was merged with the layers of a CNN. That was quite
% creative. Other ideas to bring recurrency into the network
% architecture were tried (TreeQN).

% 30 (VIN), 31 (VProp), 32 (TreeQN, ATreeC), 33 (ConvLSTM), 34
% (MCTSnets)

% Network as planner. Using a multi-layer CNN as a multi-step value iteration
% network.

% Trainable searcher.

%\subsubsection{Value Iteration Network}

%{\bf Iteration by layers}
{\bf Value Iteration Networks} (VIN) are introduced by Tamar et
al. in~\citep{tamar2016value}, see also~\citep{niu2018generalized}. A VIN is a
differentiable multi-layer network that is used to perform the steps
of a simple planning algorithm.
%VINs are
%based on the obervation  that
% Policies are reactive, in the sense that for each state a single respons
% action is learned, just like in supervised learning.
% Sequential
% decision problems require planning, or multi-step solutions. VIN is
% concerned with learning this planning behavior. A VIN can learn to
% plan, it can solve route planning in a grid-world maze, and  it can  generalize
% over such  mazes, to learn
% how to traverse unseen mazes (something that ordinary CNNs have
% difficulty with). This is a kind of transfer learning where some kind
% of abstract essence of maze-navigation is learned.
% %
%
% {\em Planner:}
The core  idea it that value
iteration (VI, see Algorithm~\ref{lst:vi}) or step-by-step planning
can be implemented by a multi-layer convolutional
network: each layer does a step of lookahead. The VI iterations are rolled-out in
the network layers $Q$ with $A$ channels.
% {\em Model:}
Through backpropagation the model learns the value
iteration parameters. The aim is to learn a general model, that can
navigate in unseen environments.
VIN learns  a fully differentiable
planning algorithm. % In order to be able to learn hard problems, curriculum learning
% is used: start to train with easy problem instances, before hard
% instances are tried. Curriculum learning~\citep{bengio2009curriculum}
% is also used in self-play (see Section~\ref{sec:selfplay}).
%
%{\em Rationale: link with other approach}
The idea of planning by gradient descent exists for some
time, several authors explored learning approximations of dynamics in
neural
networks~\citep{kelley1960gradient,schmidhuber1990line,ilin2007efficient}. 
%
%{\em Performance:}
VIN can be used for discrete and continuous path planning, and has
been tried in  grid world problems and natural language
tasks. VIN has achieved generalization of finding shortest paths in unseen mazes.

However, a limitation of VIN is that the number of
layers of the CNN restricts the number of planning steps, restricting
VINs  to small and low-dimensional domains. Schleich et 
al.~\citep{schleich2019value} extend VINs by adding abstraction, and
Srinivas et al.~\citep{srinivas2018universal} introduce universal planning networks, UPN,
which generalize to modified robot morphologies.
%
%
%\subsubsection{Value Propagation}
% {\bf Hierarchical iteration by layers}
% {\em Name:}
VProp, or {\bf Value Propagation}~\citep{nardelli2018value} is
another attempt at creating  generalizable planners
% {\em Rationale: link with other approach}
inspired by VIN.
% {\em Problem:}
By using a hierarchical structure VProp has the 
ability to generalize to larger map sizes and dynamic
environments. % {\em Model:} {\em Planner:}
VProp not only learns to plan and navigate in dynamic
environments, but  their hierarchical structure provides a way to
generalize to navigation tasks where the required planning horizon
and the size of the map are much larger than the ones seen at
training time.
%
%
% {\em Performance:}
VProp is evaluated on grid-worlds and also on dynamic environments and
on a  navigation scenario from StarCraft.

%\subsubsection{TreeQN and ATreeC}
%{\bf Tree shaped network}
%{\em Problem:}
A  different approach is taken in TreeQN/ATreeC. Again, the
aim is to create {\bf differentiable tree planning} for
deep reinforcement learning~\citep{farquhar2018treeqn}. % {\em Name:}
As VIN, TreeQN
is focused on combining planning and deep reinforcement
learning.
% {\em Rationale: link with other approach}
Unlike VIN, however, TreeQN does so by incorporating a
recursive tree structure in the network.
% {\em Model:}  
It models an MDP by
incorporating an explicit encoder function, a transition function, a
reward function, a value function, and a backup function (see also
latent models in the next subsection).
% {\em Planner:}
In this way,
it aims to achieve the same goal as VIN, that is, to create a
differentiable neural network architecture that is suitable for
planning.  %{\em Learning:}
TreeQN is based on DQN-value-functions, an actor-critic
variant is proposed as ATreeC.
TreeQN is a prelude to latent models methods in the next
subsection. In addition to being related to VIN, this approach is also related to
VPN~\citep{oh2017value} and the
Predictron~\citep{silver2017predictron}.
%
%{\em Performance:}
TreeQN is tried on  box pushing applications, like Sokoban, and nine
Atari games.

%\subsubsection{ConvLSTM}
%{\bf Planning by convolutional LSTMs}
%{\em Problem:}
Another approach to differentiable planning is to teach
a sequence of convolutional neural networks to exhibit planning
behavior.  % {\em Name:}
 A paper by~\citep{guez2019investigation} takes this approach. 
% {\em Rationale: link with other approach}
The paper demonstrates that a neural network architecture consisting
of modules of a {\bf convolutional network and LSTM} can learn to exhibit the behavior
of a planner. % {\em Planner:}
In this  approach the planning occurs implicitly, by the network, which
the authors call model-free planning, in contrast to the previous approaches in
which the network structure  more explicitly  resembles a
planner~\citep{farquhar2018treeqn,guez2018learning,tamar2016value}.
% {\em Model:}
In this method model-based behavior is learned with a general recurrent architecture
consisting of LSTMs and a convolutional
network~\citep{schmidhuber1990making} in the form of a stack of
ConvLSTM modules~\citep{xingjian2015convolutional}.
%
% {\em Learning:}
For the learning of the ConvLSTM modules the A3C
actor-critic approach is used~\citep{mnih2016asynchronous}.
%
%{\em Performance:}
The method is tried on  Sokoban and Boxworld~\citep{zambaldi2018relational}.
A stack of depth $D$, repreated $N$ times (time-ticks) allows the network
to plan. In harder Sokoban instances, larger capacity networks 
with larger depth performed better.
The experiments used a large number of environment steps,
future work should investigate how to achieve sample-efficiency with
this architecture.

%% \subsubsection{34. Planner As Network}
%% In addition to using an entire network to do planning, a planning
%% module can also be incorporated inside the neurons of a network.
%% This approach is taken by Learning to Search with
%% MCTSnets~\citep{guez2018learning}. MCTS  is a planning algorithm that
%% consists of four stages: expand, select, simulate, and backup. In this
%% work these foru stages are performed by the network, in a
%% differentiable manner.

%% Action space: discrete

%% Observation/Abstract: Observation

%% Learner/Discriminator:
%% Incorporates simulation-based search inside a neural network.  MCTSnet can
%% be trained.

%% Planner/Generator: MCTS

%% Approach: A neural
%% net that includes the four MCTS stages inside the neurons.

%% With small searches in  Sokoban MCTSnets outperform MCTS baselines.

\subsubsection*{Conclusion}
Planning networks  combine planning and transition
learning. They  fold the planning into the network,
making the planning process itself  
differentiable. The network then learns which  planning decisions to
make. Value Iteration Networks have shown how learning can transfer to mazes
that have not been seen before. A drawback
is that due to the marriage of problem size and network
topology the approach has  been limited to smaller sizes, something
that subsequent methods have tried to reduce.
One of these approaches
is TreeQN, which uses multiple smaller models and a tree-structured network. The related Predictron
architecture~\citep{silver2017predictron} also learns 
planning end-to-end, and is applicable to different kinds and sizes
of problems.

The Predictron uses abstract models, and will be
discussed in the next subsection.

\subsubsection{End-to-End Planning/Transitions with Latent Models}
We will now discuss latent model approaches in end-to-end learning of
planning and transitions.

%
%% Also RNN
%
%% 2 (SimPle), 3 (PlaNet), 5 (Dreamer), 6 (Rec World), 7 (DiscrAuto), 8
%% (GATS), 16 (VPN), 23 (SOLAR)
%
%\todo{Abstract Models is missing a red line. Is there a development
%  in the approaches? first world, then latent value, local latent, then
%  pixel2control, backprop imagination, e2e planning, zero-shot latent}
%\subsubsection{Imagination Augmented Agents} % Dit is nogal
%geavanceerd en moet naar latent models
%
%{\bf imagination by latent representation. planner, controller,
%  environment, memory. sokoban}
%{\em Name:}
The first abstract imagination-based approach that we discuss is
% \todo{Deze alinea moet nog verbeterd en ingedikt worden}
{\bf Imagination-Augmented Agent}, or I2A,
by~\citep{pascanu2017learning,racaniere2017imagination,buesing2018learning}.
A problem of model-based algorithms is
the sensitivity of the planning to model imperfections. % In
% high-dimensional domains with deep high-capacity function
% approximators,  the algorithms that perform deep lookahead suffer.
%
%{\em Problem:}
% Given that models are imperfect, 
%An important research question is how to 
% adapt  planning strategies to deal with model imperfections.  
I2A deals with these imperfections by introducing a latent model, that learns
to interpret internal simulations and adapt a strategy to the current
state.
% {\em Model:}
I2A uses latent models of the
environment, based on~\citep{chiappa2017recurrent,buesing2018learning}. The core
architectural feature of I2A is an environment model, a recurrent
architecture 
trained unsupervised from agent trajectories.
I2A has four elements that together constitute the abstract model:
(1) It has a manager that constructs a plan, which can be implemented with
a CNN. (2) It has a controller that creates an action policy. (3) It has an
environment model to do imagination. (4) Finally, it has a memory, which
can be implemented with an LSTM~\citep{pascanu2017learning}. 
%
%
% {\em Rationale: link with other approach}
%In contrast to
%the original work on imagination by Sutton, I2A is used
%in a deep learning context. 
% {\em Planner:}
I2A uses
a manager or meta-controller to choose between 
rolling out actions in the environment or by imagination (see~\citep{hamrick2017metacontrol}).
This allows  the use of models which only coarsely capture the environmental dynamics,
even when those dynamics are not perfect.  
%
% {\em Learning:}
The I2A network uses a
recurrent architecture in which a CNN is trained from agent
trajectories with A3C~\citep{mnih2016asynchronous}.
I2A  achieves success with little data and imperfect models, optimizing point-estimates of
the expected Q-values of the actions in a discrete action space. % {\em Performance:}
I2A is applied to Sokoban and
Mini-Pacman by~\citep{racaniere2017imagination,buesing2018learning}.
Performance is compared favorably to  model-free and planning
algorithms (MCTS).
Pascanu et al. apply the approach on a maze
and a spaceship task~\citep{pascanu2017learning}.
%An implementation of I2A with links to related works is available at \url{https://github.com/higgsfield/Imagination-Augmented-Agents}. 

% \todo{Add links to code implementations}

% \subsubsection{Predictron}
%
%{\bf abstract end-to-end planning and learning}
%{\em Name:}
%They introduce the {\bf Predictron}
%architecture~\citep{silver2017predictron}.
%, in which abstract models are  combined with planning.
%
%{\em Problem:}
%The goal of Predictron is to learn models that are more effective for use
%in planning than conventional model-based approaches, and that can, to
%this end, be trained end-to-end.
Planning networks (VIN)  combine
planning and learning end-to-end. A limitation of VIN is
that the tight connection between problem domain, iteration algorithm,
and network architecture limited the applicability to small grid
world problem. The {\bf  Predictron} introduces an abstract model to remove
this limitation. The  Predictron was introduced by Silver et al.\ and 
combines end-to-end planning and model learning~\citep{silver2017predictron}. 
%, just as the planning networks that we
% saw previsously.
% Planning and learning are integrated
% into one end-to-end   training procedure.
As with~\citep{oh2017value}, the model is an abstract model that  consists
of four components: a representation model, a next-state model, a reward model, and a
discount model. All models are differentiable.
%
%{\em Rationale: link with other approach}
%In principle an abstract model could generalize to many different
%prediction tasks.
% {\em Model:}
The goal of the abstract model in Predictron is to
facilitate value prediction (not state prediction) or prediction of
pseudo-reward functions that can encode special events, 
such as ``staying alive'' or ``reaching the next room.''
%
% {\em Planner:}
The planning part rolls forward its internal model $k$ steps. 
% {\em Learning:}
As in the Dyna architecture, imagined forward steps can be combined
with samples from the actual environment, combining model-free and
model-based updates.
%
%{\em Performance:}
The Predictron has been applied to procedurally generated mazes and a
simulated pool domain. In both cases it out-performed model-free
algorithms. 

% \subsubsection{World Models}
%{\bf vision, memory, control as latent. car racing}
%{\em Rationale: link with other approach}
Latent models of the dynamics of the environment can also be viewed as
{\bf World Models}, a term used by~\citep{ha2018recurrent,ha2018world}.
% {\em  Name:}
World Models are inspired
by the manner in which humans are thought to contruct a mental model of
the world in which we live. World Models are
generative recurrent neural networks that are trained unsupervised to
generate states for simulation using a variational autoencoder and a
recurrent network. They  learn a compressed spatial and temporal
representation of the environment. By using features extracted from the World Model as inputs to
the agent, a compact and simple policy can be trained to solve a
task, and planning occurs in the compressed or simplified world.
%
% {\em Model:}
For a visual environment, World Models consist of a vision model, a
memory model, and a controller. The vision model is often trained
unsupervised with a variational autoencoder. The memory model is
approximated with a mixture density network of a Gaussian
distribution (MDN-RNN)~\citep{bishop1994mixture,graves2013generating}.
%
% {\em Planner:}
The
controller model is a linear model that maps directly to actions. It
uses the CMA-ES Evolutionary Strategy for optimizing Controller models. 
Rollouts in World Models are also called {\em dreams}, to contrast
them with samples from the real environment.
%{\em Learning:}
With World Models a policy can in principle even be trained completely
inside the dream, using imagination only inside the World Model, to
test it out later in the actual environment. 
%
%{\em Performance:}
World Models have been applied experimentally to VizDoom tasks such as
Car Racing~\citep{kempka2016vizdoom}.

%% \subsubsection{7. Discrete Autoencoders}
%% Discrete autoencoders for sequence
%% models~\citep{kaiser2018discrete}

%% Improve the representation of  sequence models by introducing an
%% autoencoder that compresses the sequence through an intermediate
%% discrete latent space. 

%% Action space: Discrete sequence. language

%% Observation/Abstract: Abstract

%% delete? not a latent model. just a semantic hashing and variational
%% autoencoder technique. not reinforcemnt learnign at all

%% Learner/Discriminator: Discrete variational autoencoders for sequence models, such as text
%% and language.

%% Planner/Generator: analyze latent codes produced by the
%% model showing how they correspond to words and phrases.

%% Approach: In order to propagate gradients through this discrete representation
%% they use semantic hashing.

%% Application: Words and phrases

%\subsubsection{MuZero}
%{\bf MCTS+network, abstract}
%{\em Name:}
Taking the development of AlphaZero further is the work on 
{\bf MuZero}~\citep{schrittwieser2019mastering}.
Board games are  well suited for model-based methods because the
transition function is given by the rules of the game. % {\em
% Problem:}
However, would it be possible to do well if the rules of the game were
{\em not} given? In
MuZero a new architecture is used to learn transition functions for a
range of different games, from  Atari to board games. MuZero learns
the transition model for all games from interaction with the
environment, with one architecture, that is able to learn different
transition models.
%
%{\em Rationale: link with other approach} 
As with the Predictron~\citep{silver2017predictron} and Value
Prediction Networks~\citep{oh2017value}, MuZero has an abstract model with different
modules: representation, dynamics, and prediction function. % {\em Model:}
The
dynamics function is a recurrent process that computes transition
latent state) and
reward.  The prediction function computes policy and value functions. 
%
%{\em Planner:}
For planning, MuZero uses a version of MCTS, without the rollouts, and
with P-UCT as selection rule, using information from the abstract
model as input for node selection.
%
%{\em Learning:} 
MuZero can be regarded as joining the Predictron with self-play.
It performs well on Atari games and on board games,
learning to play the games from scratch, after having learned the
rules of the games from scratch from  the environment.

\subsubsection*{Conclusion}
Latent models represent states with a number of smaller, latent
models, allowing planning to happen in a smaller latent
space. They are useful for explicit planning and with end-to-end
learnable planning, as 
in the
Predictron~\citep{silver2017predictron}. Latent models allow
end-to-end planning to be applied to a broader range of applications,
beyond small mazes. The Predictron
creates an abstract planning network, in the spirit of, but without
the limitations of, Value Iteration
Networks~\citep{tamar2016value}. The World Models interpretation links
latent models to the way in which humans create mental models of the
world that we live in.

% Latent models can be applied with and without hybrid imagination, and with
% and without curriculum learning.
% Latent models are an innovative idea that improve both the accuracy
% of models
% and the sample efficiency.

After this detailed survey of model-based methods---the agents---it is time to
discuss our findings and draw conclusions. 
Before we do so, we will first look at  one of
the most important elements for reproducible reinforcement learning
research, the benchmark.

\begin{figure}[t]
\begin{center}
\includegraphics[width=6cm]{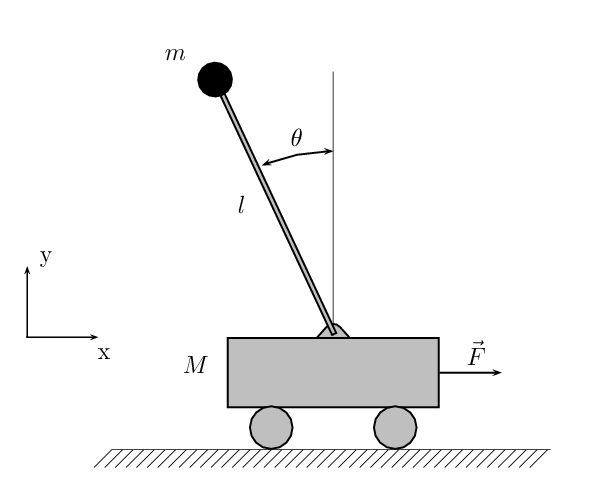}
\caption{Cartpole Pendulum~\citep{sutton2018introduction}}\label{fig:cartpole}
\end{center}
\end{figure}

\section{Benchmarks}\label{sec:bench}
%% After having discussed  different algorithms for deep model-based reinforcement learning
%% algorithms---the ---it is time to  look at the
%% algorithms themselves---the agents.

Benchmarks---the environments---play a key role in artificial
intelligence.  Without them, progress cannot be measured, and results
cannot be compared in a meaningful way. The benchmarks 
define the kind of intelligence that our artificial methods should
approach.
%Benchmarks have played a
%large role in the progress of AI.
For reinforcement learning, Mountain car and Cartpole are
well-known small problems that characterize the kind of problem to
solve (see Figure~\ref{fig:cartpole}). Chess has been
called the Drosophila of AI~\citep{landis2001aleksandr}. In addition to Mountain
car and chess a series of benchmark applications have been used 
to measure progress of artificially intelligent methods. Some of the benchmarks are
well-known and have been driving progress. In image recognition,  the
ImageNet sequence of competitions has 
stimulated  great progress~\citep{fei2009imagenet,krizhevsky2012imagenet,guo2016deep}.
The current focus on 
reproducibility in reinforcement learning emphasizes the importance of
benchmarks~\citep{henderson2017deep,islam2017reproducibility,khetarpal2018re}.

Most papers that introduce new model-based reinforcement learning
algorithms perform some form of experimental evaluation of the
algorithm. Still, since  papers use
different versions and hyper-parameter settings, comparing algorithm
performance remains difficult in practice. A recent benchmarking study compared the
performance of 14 algorithms, and some baseline algorithms on a number
of MuJoCo~\citep{todorov2012mujoco} robotics
benchmarks~\citep{wang2019benchmarking}. There was no clear winner. Performance of methods varied
widely from application to application. There is much room for
further improvement on many applications of model-based reinforcement
learning algorithms, and for making methods more robust.
The use of benchmarks should become more standardized to ease
meaningful performance comparisons.

We will now describe  benchmarks commonly used in
deep model-based reinforcement learning.
%For model-based deep reinforcement learning a diverse set of
%benchmarks is typically used.
We will discuss five sets of benchmarks: (1)
puzzles and mazes, (2) Atari arcade games such as Pac-Man, (3) 
board games such as Go and chess, (4) real-time strategy games such
as StarCraft, and (5) simulated robotics tasks such as
Half-Cheetah. As an aside, some 
of these benchmarks resemble challenges that children and adults
use to play and
learn new skills.

We will have a closer look at these sets of benchmarks, some with
discrete, some with continuous action spaces. % First
% we will discuss two benchmark sets that are often used for  planning, then 
% three benchmark sets for model-free  and model-based learning.

\begin{figure}[t]
  \begin{center}
    \includegraphics[width=8cm]{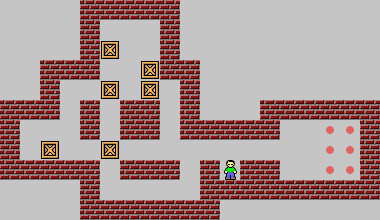}
  \end{center}
  \caption{Sokoban Puzzle~\citep{chao2013}}\label{fig:sokoban}
\end{figure}

\subsection{Mazes}
Trajectory planning algorithms are crucial in
robotics~\citep{latombe2012robot,gasparetto2015path}. There is a long
tradition of using 2D and 3D 
path-finding problems in reinforcement learning and AI.   The Taxi domain
was introduced by~\citep{dietterich2000hierarchical} in the context of
hierarchical problem solving, and box-pushing
problems such as Sokoban have been used frequently~\citep{junghanns2001sokoban,dor1999sokoban,murase1996automatic,zhou2013tabled},
see Figure~\ref{fig:sokoban}. The
action space of these puzzles and mazes is discrete. The related problems are typically  NP-hard or
PSPACE-hard~\citep{culberson1997sokoban,hearn2009games} and solving them requires
basic path and motion planning 
skills.

Small versions of the mazes can be solved exactly by planning, 
larger instances are only suitable for approximate planning or
learning methods.

Mazes can be used to test  algorithms for ``flat'' reinforcement learning
path finding problems~\citep{tamar2016value,nardelli2018value,silver2017predictron}. 
Grids and box-pushing games such as Sokoban can also feature rooms or subgoals, that may
then be used  to test algorithms for hierarchically
structured problems~\citep{farquhar2018treeqn,guez2019investigation,racaniere2017imagination,feng2020solving}.

The problems can be made more difficult  
by enlarging the grid and by inserting more obstacles. Mazes and Sokoban
grids are sometimes procedurally generated~\citep{shaker2016procedural,hendrikx2013procedural,togelius2013procedural}. The goal for the
algorithms is typically to find a solution for a grid of a certain
difficulty class, to find a shortest
solution, or to learn to solve a class of grids by training on a
different class  of grids, to test  transfer learning. 

\subsection{Board Games}
Another classic group of benchmarks for planning and  learning algorithms is board games. 

Two-person zero-sum perfect information board games such as tic tac
toe, chess,
checkers, Go, and shogi have been used since the 1950s as benchmarks in
AI. The action space of these games is discrete.  Notable achievements were in checkers, chess,
and Go, where  human world champions were defeated in 1994, 1997,
and 2016,
respectively~\citep{schaeffer1996chinook,campbell2002deep,silver2016mastering}. Other
 games are used as well as benchmarks, such as Poker~\citep{brown2018superhuman} and Diplomacy~\citep{anthony2020learning}.

The board games are typically used ``as is'' and are not changed
for different experiments (as with mazes). They are fixed benchmarks,
challenging and inspirational games where the goal is often beating 
human world champions. In model-based deep reinforcement learning they
are used for self-play methods~\citep{tesauro1995td,anthony2017thinking,schrittwieser2019mastering}.

Board games have been traditional mainstays of artificial intelligence, mostly associated
with the symbolic reasoning approach to AI. In contrast, the next benchmark is
associated with  connectionist AI.

\subsection{Atari}
%In addition to a benchmarks for traditional planning and reasoning,
Shortly after 2010 the Atari Learning Environment (ALE)~\citep{bellemare2013arcade} was introduced
for the sole purpose of evaluating reinforcement learning
algorithms on high-dimensional input, to
see if end-to-end pixel-to-joystick learning would be possible. ALE has been used widely in the
field of reinforcement learning, after impressive results such as~\citep{mnih2015human}. ALE runs on top of an emulator
for the classic 1980s Atari gaming console, and features more than 50
arcade games such as Pong, Breakout, Space Invaders, Pac-Man, and Montezuma's
Revenge.  ALE is well suited for benchmarking perception and eye-hand-coordination
type skills, less so for planning. ALE is mostly used for deep reinforcement learning
algorithms  in sensing and recognition tasks.

ALE is a popular benchmark that has been used in many model-based
papers~\citep{kaiser2019model,oh2015action,oh2017value,ha2018world,schrittwieser2019mastering}, and many model-free reinforcement learning methods.
% In 2020
% Badia et al.~\citep{badia2020agent57} reported that all 57 Atari games
% had been ``solved'' by a single agent, in the sense that the human level of
% play had been exceeded in all 57 games.
%
% ALE is mostly a challenge for interpreting high-dimensional graphical
% input. Many approaches use deep convolutional neural networks (CNN).
As with mazes, the action space is discrete and low-dimensional---9 joystick directions
and a push-button---although the input space is high-dimensional.

The ALE games are quite varied in nature. There are ``easy''
eye-hand-coordination tasks such as Pong and Breakout, and there are
more strategic level games where long periods of movement exist
without changes in score,
such as Pitfall and Montezuma's Revenge. The goal of ALE experiments
is typically to achieve a score level 
comparable to humans in as many of the 57 games as possible (which has
recently been achieved~\citep{badia2020agent57}).
After this achievement, some researchers believe that the field is
ready  for more challenging benchmarks~\citep{machado2018revisiting}.  

\subsection{Real-Time Strategy and Video Games}\label{sec:rts}
The Atari benchmarks are based on simple arcade games of 35--40 years ago,
most of which are mostly challenging for eye-hand-coordination
skills. Real-time strategy  (RTS) games and games such as
StarCraft, DOTA, and Capture the Flag provide more challenging
tasks. The strategy space is large; the state space of StarCraft has
been estimated at $10^{1685}$, much larger than chess ($10^{47}$) or
go ($10^{147}$).  Most RTS games are multi-player, non-zero-sum, imperfect information
games that also feature high-dimensional pixel input,
 reasoning,  team
collaboration, as well as  eye-hand-coordination. The action space
is stochastic and is a mix of discrete and continuous actions. A very
high degree of diversity is necessary to prevent specialization.

Despite the challenging nature, impressive achievements have been
reported recently in all three mentioned
games where human performance was matched or even exceeded~\citep{vinyals2019grandmaster,berner2019dota,jaderberg2019human}. In
these efforts deep model-based reinforcement learning is combined with
multi-agent and population based methods. These mixed approaches
achieve impressive results on RTS games; added diversity diminishes the
specialization trap of two-agent approaches~\citep{vinyals2019grandmaster}, their
approaches may combine aspects of self-play and latent
models, although  often the well-tuned combination of a number
of methods is credited with the high achievements in these games.
The mixed approaches are not listed separately in the taxonomy of the next section.

\begin{figure}[t]
  \begin{center}
    \includegraphics[width=8cm]{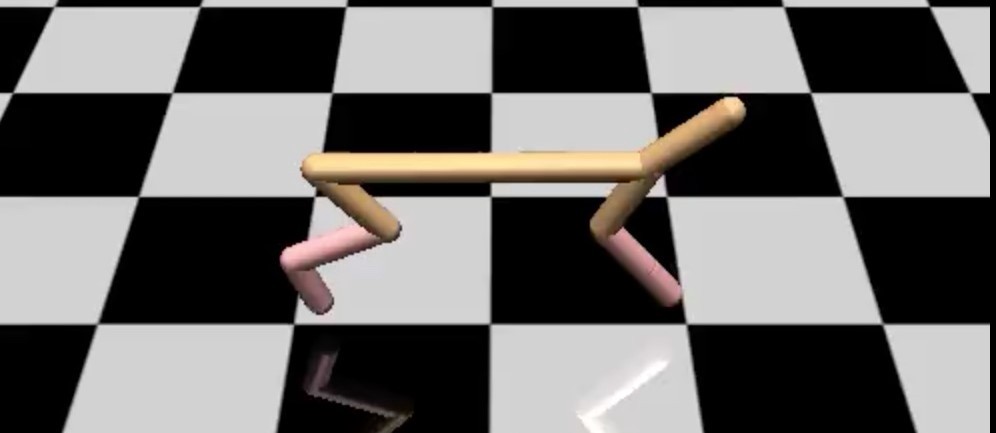}
  \end{center}
  \caption{Half-Cheetah}\label{fig:cheetah}
\end{figure}

\subsection{Robotics}
Reinforcement learning is a paradigm that is well-suited for modeling
planning and control problems in robotics. Instead of minutely programming
high-dimensional robot-movements step-by-step, reinforcement learning is used to train
behavior  more flexibly, and possibly end-to-end from camera input to
arm manipulation.

Training with real-world robots is expensive
and complicated. In robotics sample
efficiency is of great importance because of the cost of interaction with
real-world environments and the wear of physical robot arms. For this
reason virtual
environments  have been devised. Todorov et al.~\citep{todorov2012mujoco} introduced
MuJoCo, a software suite for simulated robot behavior. It is used
extensively in reinforcement learning research. Well-known benchmark
tasks are Reacher, Swimmer, Half-Cheetah, and Ant, in which the
agent's task is to teach itself the appropriate movement actions, see
Figure~\ref{fig:cheetah} for an example. Many model-based deep
reinforcement learning methods are tested on
MuJoCo~\citep{heess2015learning,chua2018deep,
  % kurutach2018model,
janner2019trust,gu2016continuous,feinberg2018model,clavera2018model,azizzadenesheli2018surprising,hafner2018learning} and other
robotics
tasks~\citep{tassa2012synthesis,levine2014learning,finn2017deep,
%  zhang2018solar,
  hafner2019dream,sekar2020planning}.
The action space of these tasks is continuous, and the emphasis in
experiments is on sample efficiency.

\subsubsection*{Conclusion}
No discussion on empirical  deep reinforcement learning is complete
without the mention of OpenAI Gym~\citep{brockman2016openai}.  Gym is a
toolkit for developing and 
comparing reinforcement learning algorithms and provides a training
environment for
reinforcement learning agents. %It supports teaching
%agents everything from walking to playing games from Pong to
%Pinball.
Gym includes interfaces to benchmark sets such as ALE and
MuJoCo. % Gym provides an easy to use Python interface, and is
% well-documented.
% Gym has lowered the barrier of entry to doing
% research in the field greatly, and has helped  the
% growth of the field.
Other software suites are
\citep{tassa2018deepmind,vinyals2017starcraft,yu2020meta,bellemare2013arcade,todorov2012mujoco}.

Baseline implementations of
many  deep reinforcement learning agent algorithms can also be found
at the Gym website \url{https://gym.openai.com}.  

Research into suitable benchmarks is active. Further interesting approaches
are Procedural Content Generation~\citep{togelius2013procedural}, 
MuJoCo Soccer~\citep{liu2019emergent}, and the Obstacle Tower Challenge~\citep{juliani2019obstacle}.

Now that we have seen the benchmarks on which our model-based methods
are tested, it is time for an in-depth discussion and outlook for
future work.

\section{Discussion and Outlook}\label{sec:dis}

% The goal of reinforcement learning is to learn the policy function
% that provides the best action in all states. Model-based reinforcement
% learning combines   a transition model with planning and
% learning, improving sample efficiency and generalization. We have
% noted  research directions that  
% work such as abstract models and self-play, and even abstract models
% in combination with self-play. It will be interesting to see if
% planning networks can be combined with other approaches.

%{\bf FUTURE: 0. tx models for seq dec prob \& transfer learning/zero
%  shot, 1. high dim works now through abstract models and 2.  curriculum learning }

%An often heard reason  for the interest in
Model-based reinforcement learning  promises  lower sample
complexity. Sutton's work on imagination, where a model is created
with environment samples that are then used to create extra imagined samples
for free, clearly suggests this aspect of model-based reinforcement
learning. The  transition model acts as a multiplier on the
amount of information that is used from each environment sample.

Another, and perhaps more important aspect, is generalization
performance. Model-based reinforcement learning builds a dynamics model of the
domain. This model can be used multiple times, for new problem
instances, but also for new problem classes. By learning the
transition and reward model, model-based reinforcement 
learning may be better at capturing the essence of a domain  than 
model-free methods. Model-based reinforcement
learning may thus be better suited for solving transfer learning problems,
and for solving long sequential decision making problems, a class of
problems that is important in real world decision making.

Classical table based approaches and Gaussian Process approaches have
been quite succesful in achieving low sample complexity for problems
of moderate
complexity~\citep{sutton2018introduction,deisenroth2013survey,kober2013reinforcement}.
However,
the topic of the current survey is  \emph{deep} models, for high 
dimensional problems with complex, non-linear, and discontinuous
functions. These application domains pose a problem for classical model-based approaches.
Since high-capacity deep neural networks  require many
samples to achieve high generalization (without  overfitting), a
challenge  in deep model-based reinforcement learning is to maintain low sample complexity.

We have  seen a
wide range of approaches that attempt this goal. % Most methods were
% overwhelmingly from recent years. 
% Recent years have seen many new methods. We identified 
% three  categories of approaches.
% % Model learning and imagination are probably closest to the original
% % (pre-deep learning) model-based
% % ideas, and abstract models are the latest and most promising
% % incarnation. Planning networks and self-play are less main-stream,
% % and the future  will learn if more can be expected from these approaches.
% A key aspect of deep model-based reinforcement learning is the balance
% between bias and performance. Despite some success with probability
% based models, single models and imagination have difficulty in achieving low sample
% complexity and high performance in high-dimensional end-to-end
% model-based reinforcement learning. 
% Curriculum learning, uncertainty modeling, latent models, and
% end-to-end model learning and planning, 
% have been able to achieve this goal.
%
Let us now discuss and summarize the benchmarks, the approaches for
deep models, and possible future work.

\subsection{Benchmarking}

Benchmarks are the lifeblood of AI. We must test our algorithms to
know if they exhibit intelligent behavior.
%Fortunately,
%there is a wide spectrum of benchmarks to choose from, from mazes,
%through strategy games, to robot manipulation tasks, both simulated
%and real.
Many
of the benchmarks allow difficult decision making situations to be
modeled. Two-person games allow modeling of competition. In real
world decision making, collaboration and negotiation are also
important. Real-time strategy games allow collaboration, competition
and negotiation to be modelled, and multi-agent and hierarchical
algorithms are being developed for these decision making
situations~\citep{jaderberg2019human,kulkarni2016hierarchical,makar2001hierarchical}.

Unfortunately, the wealth of choice  in benchmarks makes it
difficult to compare results that are reported in the
literature. Not all authors publish their code. We typically
need to rerun experiments with identical benchmarks to compare
algorithms conclusively. Outcomes often differ from  the original
works, also because not all hyperparameter settings are always clear
and implementation details may
differ~\citep{henderson2017deep,islam2017reproducibility,wang2019benchmarking}. Authors
should publish their code if our field wants to make progress.
The recent attention for reproducibility in  deep reinforcement learning is
helping the field move in the right direction. Many papers now
publish their code and the hyperparameter settings that were used in the reported
experiments.

\subsection{Curriculum Learning and Latent Models} % +curriculum,
% ?abstract
Model-based reinforcement learning works well when the transition and
reward models are {\bf given} by the rules of the problem. We have seen how
perfect models in games such as chess and Go allow deep and accurate
planning. Systems were constructed~\citep{tesauro1995td,silver2018general,anthony2017thinking} where
curriculum learning facilitated tabula rasa self-learning of highly
complex games of
strategy; see also~\citep{bengio2009curriculum,plaat2020learning,narvekar2020curriculum}.
%
% in en efficient curriculum learning 
% We start with methods for problems where transitions are provided by
% rules of the game. The transitions can then be used with classic planning algorithms to
% learn the optimal policy and value function. 
% By using the learning agent as its own environment, self-play creates
% a curriculum learning setting in which the agent is exposed to a
% sequence of increasingly hard learning tasks. The planner uses the
% network to generate sample positions, with game outcomes, that are
% used to improve the network, which is then used to generate better samples,
% etc.  In this way a learning curriculum is generated that  starts with easy
% tasks, progressing to harder tasks, learning hard tasks quickly
% and
% efficiently~\citep{bengio2009curriculum,silver2017mastering,narvekar2020curriculum}. Impressive  
% results have been achieved learning to play Backgammon, chess and Go
% from scratch---tabula rasa---to World-Champion level.
% It
% is straightforward to combine agent and environment in a two-agent
% setting, and self-play has been used frequently in two-agent
% games~\citep{plaat2020learning}. 
The success of self-play has led to interest in curriculum learning
in single-agent
problems~\citep{feng2020solving,doan2019line,duan2016rl,laterre2018ranked}.  

When the rules are not given, they might be {\bf learned}, to create a
transition model. Unfortunately, the planning accuracy of learned models is less than
perfect. We have seen  efforts with Gaussian Processes and ensembles to
improve model quality, and efforts with local planning and 
Model-Predictive Control, to limit the damage of imperfections in the
transition model. We have also discussed, at length, latent,
abstract, models, where for each of the functions of the Markov
Decision Process a separate sub-module is learned.  Latent models
achieve better accuracy with explicit
planning, as planning occurs in latent
space~\citep{oh2017value,hafner2019dream,kaiser2019model}.  

\begin{figure}
  \begin{center}
  \begin{tikzpicture}[->,scale=1,font=\footnotesize,
  desc/.style={ 
		scale=1.0,
		rectangle,
		rounded corners,
                draw=black,
                very thick
		},
  descg/.style={
		scale=1.0,
		rectangle,
		rounded corners,
                draw=black!55!green,
                very thick
		},
  descy/.style={
		scale=1.0,
		rectangle,
		rounded corners,
                draw=black!30!yellow,
                very thick
		},
  descr/.style={
		scale=1.0,
		rectangle,
		rounded corners,
                draw=black!25!red,
                very thick
		}]
  \node[desc] (mbdrl) at (180:1cm)  {\bf MBDRL};
  \node[descg] (given) at (145:1.5cm)  {\bf given model};
  \node[descg] (td) at (138:2.5cm) {TDgammon};
  \node[descg] (exit) at (138:4cm) {ExIt};
  \node[descg] (soko) at (125:4.8cm) {Single Cur};
  \node[descg] (ag) at (115:3.2cm) {AlphaGo};
  \node[descg] (az) at (90:4cm) {AlphaZero};
  
  \node[descy] (end2end) at (45:1.4cm) {\bf end2end};
  \node[descy,dashed] (pred) at (40:4.5cm) {Predictron};
  \node[descy] (vprop) at (85:2.8cm) {VProp};
  \node[descy,dashed] (treeqn) at (60:3.3cm) {TreeQN};
  \node[descy,dashed] (i2a) at (10:3cm) {I2A};
  \node[descy,dashed] (mu) at (50:5.3cm) {MuZero};
  \node[descy,dashed] (world) at (350:1.7cm) {World model};
  \node[descy] (convlstm) at (80:2cm) {ConvLSTM};
  \node[descy] (vin) at (30:3cm) {VIN};

  \node[descr] (learned) at (240:1cm) {\bf learned model};
  \node[descr] (expl) at (215:3cm) {explicit plan};
  \node[descr] (pilco) at (190:2.5cm) {PILCO};
  \node[descr] (ilqg) at (192:3.8cm) {iLQG};
  \node[descr] (gps) at (230:3.7cm) {GPS};
  \node[descr] (fore) at (215:4.5cm) {Foresight};
  \node[descr] (svg) at (242:4.2cm) {SVG};
  \node[descr] (pets) at (230:4.8cm) {PETS};
  \node[descr,dashed] (planet) at (265:4.5cm) {PlaNet};
  \node[descr,dashed] (dreamer) at (245:4.9cm) {Dreamer};
  \node[descr,dashed] (plan2expl) at (230:5.8cm) {Plan2Expl};

  \node[descr] (imag) at (325:3.2cm) {imagination};
  \node[descr,dashed] (vpn) at (310:5.7cm) {VPN};
  \node[descr] (local) at (250:2cm) {Local};
  \node[descr] (mve) at (250:3cm) {MVE};
  \node[descr] (pol) at (270:3.4cm) {PolicyOptim};
  \node[descr] (meta) at (288:2.6cm) {Meta};
  \node[descr] (gats) at (295:3.4cm) {GATS};
  \node[descr,dashed] (simple) at (292:4.3cm) {SimPLe};
  \node[descr,dashed] (video) at (310:4.4cm) {VideoPred};

   %\draw[thick]  (60:1.5cm)  arc   (60:-20:1.5cm)  node[auto] {acting} ;
  % \draw[thick] (-40:1.5cm) arc  (-40:-140:1.5cm) node[auto] {model learning} ;
 %  \draw[thick]  (200:1.5cm) arc (200:120:1.5cm)  node[auto] {planning};
%   \draw[thick]  (-28:1cm) arc  (240:160:1.4cm) node[auto] {direct
%     RL};

   \draw[black!55!green,->,very thick]  (mbdrl) to  (given);
   \draw[black!55!green,->,very thick]  (given) to  (td);
   \draw[black!55!green,->,very thick]  (td) to  (ag);
   \draw[black!55!green,->,very thick]  (td) to  (exit);
   \draw[black!55!green,->,very thick]  (ag) to  (az);
   \draw[black!55!green,->,very thick]  (az) to  (soko);
   \draw[black!25!red,->,very thick]  (mbdrl) to  (learned);
   \draw[black!25!red,->,very thick]  (learned) to  (expl);
   \draw[black!25!red,->,very thick]  (learned) to  (imag);
   \draw[black!25!red,->,very thick]  (learned) to  (end2end);
   \draw[black!30!yellow,->,very thick]  (end2end) to  (vin);
   \draw[black!30!yellow,->,very thick,dashed]  (end2end) to  (world);
   \draw[black!30!yellow,->,very thick]  (end2end) to  (convlstm);
   \draw[black!30!yellow,->,very thick,dashed]  (vin) to  (i2a);
   \draw[black!30!yellow,->,very thick,dashed]  (vin) to  (pred);
   \draw[black!30!yellow,->,very thick]  (vin) to  (vprop);
   \draw[black!30!yellow,->,very thick,dashed]  (vin) to  (treeqn);
   \draw[black!30!yellow,->,very thick,dashed]  (pred) to  (mu);
   \draw[black!55!green,->,very thick,dashed]  (az) to  (mu);
   \draw[black!25!red,->,very thick,dashed]  (vpn) to  (planet);
   \draw[black!25!red,->,very thick,dashed]  (vpn) to  (pred);
   \draw[black!25!red,->,very thick]  (expl) to  (gps);
   \draw[black!25!red,->,very thick]  (expl) to  (ilqg);
   \draw[black!25!red,->,very thick]  (expl) to  (pilco);
   \draw[black!25!red,->,very thick,dashed]  (expl) to  (planet);
   \draw[black!25!red,->,very thick,dashed]  (planet) to  (dreamer);
   \draw[black!25!red,->,very thick,dashed]  (dreamer) to  (plan2expl);
   \draw[black!25!red,->,very thick]  (gps) to  (fore);   
   \draw[black!25!red,->,very thick]  (gps) to  (svg);   
   \draw[black!25!red,->,very thick]  (gps) to  (pets);   
   \draw[black!25!red,->,very thick,dashed]  (imag) to  (i2a);   
   \draw[black!25!red,->,very thick]  (local) to  (pol);   
   \draw[black!25!red,->,very thick]  (local) to  (mve);   
   \draw[black!25!red,->,very thick]  (local) to  (meta);   
   \draw[black!25!red,->,very thick]  (imag) to  (local);   
   \draw[black!25!red,->,very thick]  (imag) to  (gats);   
   \draw[black!25!red,->,very thick,dashed]  (imag) to  (video);   
   \draw[black!25!red,->,very thick,dashed]  (imag) to  (simple);   
   \draw[black!25!red,->,very thick,dashed]  (video) to  (vpn);

\end{tikzpicture}
 \caption{Influence of Model-Based Deep Reinforcement Learning
   Approaches. Green: given transitions/explicit planning; Red: learned
   transitions/explicit planning; Yellow: end-to-end
   transitions and planning. Dashed: latent models. }\label{fig:influence}
 \end{center}
 \end{figure}
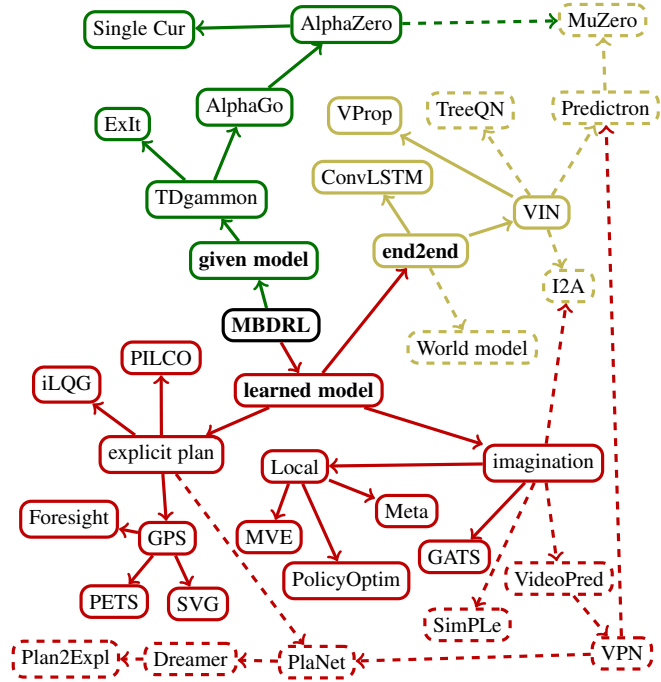

The work on Value Iteration Networks~\citep{tamar2016value} inspired
{\bf end-to-end} learning,
where both the  transition model and the planning algorithm are
learned, end-to-end. Combined with latent models (or World Models~\citep{ha2018world}) impressive results
where achieved~\citep{silver2017predictron}, and model/planning
accuracy was improved to the extent that tabula rasa curriculum
self-learning was achieved, in Muzero~\citep{schrittwieser2019mastering} for 
both chess, Go, and Atari games. End-to-end learning and latent models
together allowed the circle to be closed, achieving curriculum
learning self-play also for problems where the rules are not
given. See Figure~\ref{fig:influence} for relations between the
different approaches of this survey. The main categories are
color-coded, latent methods are dashed.

% In two-agent games the transition model and the reward model are given
% by the rules of the game, and there is no need to learn them from
% sampling the environment, when the game is known.
% The combination of self-play with latent models in
% MuZero~\citep{schrittwieser2019mastering}, however, has shown that
%  transition models can be learned from unknown games as different as
% chess and Atari. 
Optimizing directly in a latent space has  been
successful  with a generative adversarial
network~\citep{volz2018evolving}.  World Models  are linked to
neuroevolution by~\citep{risi2019deep}. For future work, the combination of curriculum
learning, ensembles, and latent models appears quite fruitful. 
Self-play has been used to achieve further success in other
challenging games, such as StarCraft~\citep{vinyals2019grandmaster}, DOTA~\citep{berner2019dota},
Capture the Flag~\citep{jaderberg2019human}, and Poker~\citep{brown2019superhuman}. In multi-agent real-time strategy
games the aspect of collaboration and teams is  important, and self-play
model-based reinforcement learning has been combined with multi-agent, hierarchical,
and population based methods. 

In future work,  more 
research is needed to explore the potential of (end-to-end) planning
with latent representational models more fully for larger problems,
for transfer learning, and 
for cooperative problems.  More research is needed in uncertainty-aware neural networks.
For such challenging problems as real-time strategy
and other video
games,  more combinations of deep model-based reinforcement learning with
multi-agent and population based methods can be
expected~\citep{vinyals2019grandmaster,risi2020chess,back1996evolutionary}.

In end-to-end planning and  learning, latent models introduce 
differentiable versions of more and more classical algorithms.
For example, World Models~\citep{ha2018world} have a trio of
Vision, Memory, Control models, reminding us of the Model,
View, Controller design pattern~\citep{gamma2009design}, or even of classical computer
architecture elements such as ALU,
RAM, and Control
Unit~\citep{hennessy2011computer}. Future work will
show if differentiable algorithms will be found for even more classical algorithms.

\section{Conclusion}\label{sec:con}
Deep learning has revolutionized  reinforcement
learning. The new methods allow us to approach more complicated
problems than before. Control and decision making tasks involving
high dimensional visual input  come within reach.

In principle, model-based methods offer the advantage of lower sample
complexity over model-free methods, because of their transition model.
However, traditional methods, such as Gaussian Processes, that work well on
moderately complex problems with few samples, do not perform
well on high-dimensional problems. High-capacity models have high
sample complexity, and finding methods that generalize well with low
sample complexity has been difficult.

In the last five years many new methods have been devised,
and great success has been achieved in model-free and in model-based deep
reinforcement learning. This survey has summarized the main ideas of
recent papers in three 
approaches. For more and more aspects of model-based reinforcement learning
algorithms differentiable methods appear. Latent models condense
complex problems into compact latent representations that are easier
to learn. End-to-end curriculum learning of latent planning and learning models
has been achieved.

In the discussion we mentioned  open problems for each of the
approaches, where we expect worthwhile  future work will
occur, such as curriculum learning,
uncertainty modeling and transfer learning by latent models.
Benchmarks are important to test progress, and more benchmarks of
latent models can be expected. Curriculum learning has shown how
complex problems can be learned relatively quickly from
scratch, and latent models allow  planning in efficient latent
spaces. Impressive results have been  
reported; future work can be expected in transfer learning with 
latent models, and the interplay of curriculum learning with
(generative)  latent
models, in combination  with end-to-end learning of larger problems.

Benchmarks in the field have also had to keep up.
Benchmarks have progressed from single-agent grid worlds to
multi-agent Real-time strategy games and  
complicated camera-arm robotics manipulation tasks.
Reproducibility and benchmarking studies are of great importance for
real progress. In real-time strategy games model-based methods are 
being combined with multi-agent, hierarchical  and evolutionary
approaches, allowing the study of collaboration, competation and
negotiation.

Model-based deep reinforcement learning is a vibrant field of
AI with a long history before deep learning. The field is blessed with
a high degree of activity, 
an open culture, clear benchmarks,  shared
code-bases~\citep{brockman2016openai,vinyals2017starcraft,tassa2018deepmind}
and a quick turnaround of 
ideas. We hope that this survey will   lower 
the barrier of entry even further.

\section*{Acknowledgments}
We thank the members of the Leiden Reinforcement
Learning Group, and especially Thomas Moerland and Mike Huisman,
for many
discussions and insights.

{\footnotesize
\bibliographystyle{plainnat}
\bibliography{\string~/Dropbox/BibTex/plaat}
}

\end{document}